\documentclass[10pt]{article} 
\usepackage[accepted]{tmlr}

\usepackage{amsmath,amsfonts,bm}

\def\eqref#1{equation~\ref{#1}}

\def\1{\bm{1}}

\DeclareMathAlphabet{\mathsfit}{\encodingdefault}{\sfdefault}{m}{sl}
\SetMathAlphabet{\mathsfit}{bold}{\encodingdefault}{\sfdefault}{bx}{n}

\newcommand{\R}{\mathbb{R}}

\newcommand{\abs}[1]{\left| #1 \right|}
\newcommand{\Ex}[1]{\mathbb{E}\left[#1\right]}
\newcommand{\Prob}[1]{\mathbb{P}\left[#1\right]}

\usepackage{color}

\usepackage{hyperref}
\usepackage{url}
\usepackage{graphicx}
\usepackage{booktabs}
\usepackage{mathtools}
\usepackage{amsthm}
\usepackage{listings}
\usepackage{placeins}
\usepackage{siunitx}
\usepackage{algorithm}
\usepackage{algorithmic}
\usepackage{eqparbox}
\usepackage{xcolor}
\usepackage{todonotes}

\title{A Faster Generalized Two-Stage Approximate Top-K}

\author{\name Yashas Samaga \email syashas@cs.washington.edu \\
      \addr University of Washington, Seattle \\
      (work done while at Google DeepMind)
      \ANDauthor
      \name Varun Yerram \email y.varun@nyu.edu \\
      \addr New York University \\
      (work done while at Google DeepMind)
      \ANDauthor
      \name Spandana Raj Babbula \email sbabbula@google.com \\
      \addr Google DeepMind
      \ANDauthor
      \name Prateek Jain \email prajain@google.com \\
      \addr Google DeepMind
      \ANDauthor
      \name Praneeth Netrapalli \email pnetrapalli@google.com \\
      \addr Google DeepMind
}

\def\month{05}  
\def\year{2026} 
\def\openreview{\url{https://openreview.net/forum?id=izqZ1Crpjz}} 

\begin{document}

\maketitle

\begin{abstract}
We consider the Top-$K$ selection problem, which aims to identify the largest $K$ elements in an array. Top-$K$ selection arises in many machine learning algorithms and often becomes a bottleneck on accelerators, which are optimized for dense matrix multiplications. To address this problem, \citet{chern2022tpuknnknearestneighbor} proposed a fast two-stage \textit{approximate} Top-$K$ algorithm that: (i) partitions the input array into equal-sized chunks and selects the top-$1$ element from each partition; and (ii) sorts the resulting \textit{smaller subset} and returns the top $K$ elements. In this paper, we generalize the first stage so that each partition selects the top $K'$ elements (for $1 \leq K' \leq K$). Our contributions include: (i) an expression for the expected recall of this generalized algorithm under random partitioning, and a demonstration that choosing $K' > 1$ with \textit{fewer partitions} in the first stage more effectively reduces the input size to the second stage while maintaining the same expected recall as the original algorithm; (ii) a bound on the expected recall of the original algorithm as a function of the algorithm parameters that is provably tighter by a factor of $2$ than the bound reported by \citet{chern2022tpuknnknearestneighbor}; and (iii) an implementation of our algorithm on Cloud TPUv5e that achieves approximately an order of magnitude speedup over the original algorithm without sacrificing recall.
\end{abstract}

\section{Introduction}
Identifying the top-K elements in an array is an essential building block of many algorithms. Beyond its common applications in Maximum Inner Product Search (MIPS) or K-Nearest-Neighbors (KNN)~\citep{chern2022tpuknnknearestneighbor, li2023recentdevelopmentsrecommendersystems}, it has recently become important for optimizing training and inference of large foundation models. Large language models (LLMs) use the Top-$K$ operation to exploit the sparsity in several components, including classification logits \citep{l2024hirehighrecallapproximate}, 
MLP blocks~\citep{liu2023dejavucontextualsparsity,l2024hirehighrecallapproximate, alizadeh2024llmflashefficientlarge}, 
attention mechanisms \citep{roy2021efficient,madaan2023treeformerdensegradienttrees, DBLP:journals/corr/abs-2512-16391, synk2025exploitingsparsitylongcontext} and KVCache compression \citep{behnam2025rocketkv,yang2025tidaldecode, tang2024quest}. It is also used in 
retrieval augmented generation~\citep{lewis2021retrievalaugmentedgenerationknowledgeintensivenlp, borgeaud2022improvinglanguagemodelsretrieving}, 
sampling tokens~\citep{shen2024superposeddecodingmultiplegenerations}, mixture-of-experts~\citep{dai-etal-2024-deepseekmoe, he2024mixturemillionexperts}, and accelerating distributed training~\citep{shi2019understandingtopksparsificationdistributed, ruan2023adaptivetopksgdcommunicationefficient}.

Given the scale of these models, their training and inference are typically performed on accelerators such as TPUs and GPUs. However, computing Top-$K$ on these devices can be expensive. On TPUv5e and A100, finding the top-2\% of the hidden activations of Gemma~2 9B's \citep{gemmateam2024gemmaopenmodelsbased} feedforward blocks\footnote{This involves an einsum contraction between a 3D tensor of shape \texttt{[batch\_size, seqlen, model\_dims]} and a 2D weight matrix of shape \texttt{[model\_dims, hidden\_dims]}, contracting along the \texttt{model\_dims} axis (i.e., \texttt{"bsm,mh~->~bsh"}). Top-K is then applied along the \texttt{hidden\_dims} axis.} during training using \texttt{jax.lax.top\_k} takes 27$\times$ and 4.8$\times$ longer, respectively, than the corresponding matrix multiplication that generated those activations. Ideally, computing Top-$K$ should be negligible compared to the matrix multiplications.

As a workaround, foundation models increasingly use \textit{approximate} Top-K algorithms, which are generally robust to these approximations~\citep{l2024hirehighrecallapproximate, key2024approximate}.

\citet{chern2022tpuknnknearestneighbor} introduced a hardware-friendly approximate Top-$K$ algorithm that operates in \textit{two stages}. In the first stage, the input array is divided into a fixed number of equal-sized buckets, and the top-$1$ element from each bucket is selected. In the second stage, these top-$1$ elements are sorted, and the first $K$ elements are returned. They quantify the approximation error in terms of \textit{expected recall} \citep{wang2014hashingsimilaritysearchsurvey}, defined as the proportion of the actual top-K elements retrieved in the first stage averaged over all permutations of the inputs. They derive a closed-form expression that relates the expected recall to the number of buckets, which is then used to \textit{choose a number of buckets} that is sufficient to maintain a \textit{user-specified average recall target}. We refer to this method as \emph{the original algorithm}.

At the time of writing, this algorithm was implemented in \texttt{jax.lax.approx\_max\_k} for TPUs. In the earlier example of finding the top-2\% of the hidden activations, this algorithm with a recall target of 95\% still takes 9$\times$ more time than the matrix multiplication on TPUv5e.

As we explain in Sections~\ref{sec:background} and~\ref{sec:method}, the second stage is typically the bottleneck, since the first-stage computation is relatively inexpensive and efficiently parallelizable. In fact, in tasks that require finding the top-K elements in each column or row of a matrix product, the first stage can be ``fused"\footnote{In this context, \textit{fusion} refers to merging the top-K selection logic with the matrix multiplication logic into a single fused operation such that the top-K step can, in some cases, be performed at no additional cost. We discuss this in more detail in Section \ref{sec:background}.}~\citep{snider2023operatorfusionxlaanalysis} with the matrix multiplication, effectively hiding much of its cost. Therefore, improving the efficiency of this algorithm requires reducing the number of elements sorted in the second stage without sacrificing the expected recall, while ensuring that the first-stage computation remains inexpensive. Our main contribution is a new algorithm that achieves this, which we describe next.

We accomplish this by generalizing the first stage of the approximate Top-$K$ algorithm of~\citet{chern2022tpuknnknearestneighbor} to select top-$K'$ elements from each bucket (where $K' < K$) instead of restricting selection to the top-$1$. This increases the total number of elements sorted in the second stage to $B \cdot K'$, where $B$ is the number of buckets. However, \textit{our key result shows that for a large set of values of $K$, array size $N$ and recall targets, it is possible to reduce the number of buckets $B$ \textit{sufficiently} that the smallest number of elements to sort in the second stage ($B\cdot K'$) is achieved by some $K' > 1$  while ensuring that the first stage does not become the bottleneck.}

\textbf{Theoretically}, we derive an expression relating $K'$, $K$ and $B$ to the expected recall. Using this expression, we select the parameters $K'$ and $B$ for our algorithm that satisfy the user-specified average recall target. While the full potential is realized by choosing $K'>1$, interestingly, even for $K'=1$, the setting of~\citet{chern2022tpuknnknearestneighbor}, our bound on the number of buckets is provably a factor of $2$ tighter than theirs, which improves parameter selection and, in most cases, more than doubles the performance of the original algorithm. For $K' > 1$, the gains are even higher.

\textbf{Empirically}, we implement our generalized algorithm ($K' > 1$) for TPUs using Pallas and demonstrate an order of magnitude speedup in latency on TPUv5e chips compared to the original algorithm. We provide two implementations: (i) an unfused version that executes the two stages as two separate \textit{kernels} (background on "kernels" is in Section \ref{sec:background}); and (ii) a matmul-fused version that fuses the first stage with a matrix multiplication (background on fusion and related matmuls are in Sections~\ref{sec:background} and~\ref{sec:problem}). In the earlier example of identifying the top 2\% of hidden activations, our implementations for $K'=4$ make the Top-$K$ step 24$\times$ faster than \texttt{jax.lax.approx\_max\_k}. This reduces the time taken by approximate Top-$K$ to below that of the corresponding matrix multiplication, resulting in an overall speedup of 6.7$\times$. These localized gains translate into significant end-to-end gains; in Appendix \ref{sec:speedups_sparse_mlp_training}, we provide a holistic breakdown of the training costs of a sparse transformer block. We share the code for our TPU implementations in Appendix \ref{lst:pallas_implementation} and \ref{lst:fused_pallas_implementation}. 


\section{Background}
\label{sec:background}
\subsection{Organization of compute on accelerators}
Compute resources in most accelerators are distributed across several distinct subsystems, each specialized for different types of operations (see Figure~\ref{fig:tpuv5e_arch}). On TPUs, the vast majority of compute is concentrated in two subsystems: the Matrix Multiply Unit (MXU) and the Vector Processing Unit (VPU) \citep{9220735, tpu_system_arch}. MXUs account for more than 95\% of compute FLOPS \citep{chern2022tpuknnknearestneighbor, tpu_system_arch}; consequently, only MXU-bound programs reach peak FLOPS utilization. Similarly, on GPUs, compute resources are primarily spread across two subsystems: TensorCores for matrix multiplications and CUDA cores for scalar/vector computations. As with TPUs, most FLOPS are concentrated in TensorCores \citep{h100_specs}.


\subsection{Kernels and fusions}
\label{sec:fusion}
\begin{figure}[tbp]
  \begin{minipage}{0.48\linewidth}
    \centering
    \includegraphics[width=\linewidth]{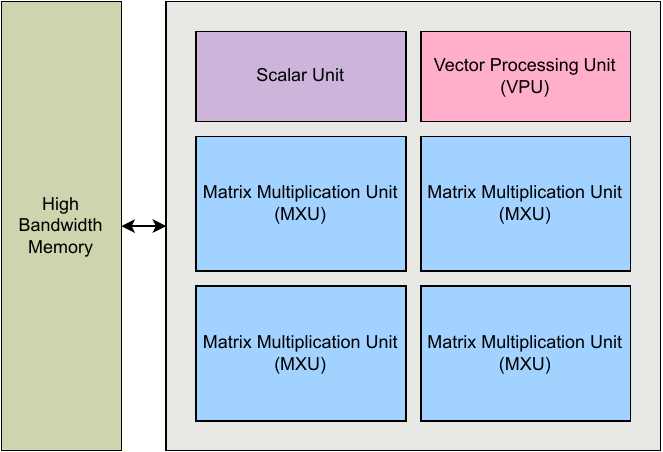}
    \caption{\textbf{Overview of TPUv5e subsystems.} The TPUv5e chip features four Matrix Multiplication Units (MXUs) dedicated to matrix-matrix multiplications along with a Vector Processing Unit (VPU) that performs general vector operations, such as activations and softmax. The chip also includes a scalar unit for calculating memory addresses, managing control flow, and performing similar tasks.}
    \label{fig:tpuv5e_arch}
  \end{minipage}
  \hfill
  \begin{minipage}{0.48\linewidth}
    \centering
    \includegraphics[width=\linewidth]{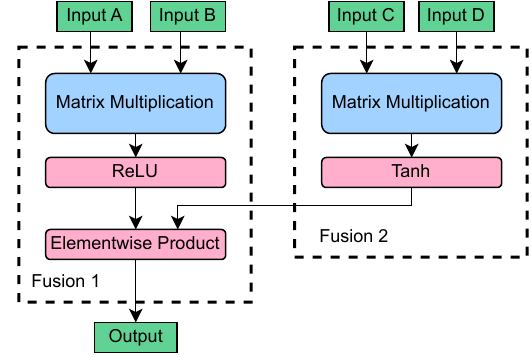}
    \caption{\textbf{Decomposing a program into smaller subprograms.} In this example, a program has been broken down into two subprograms, each containing several elementary operations. The subprograms are executed in an order that satisfies their dependencies, with Fusion~2 executing before Fusion~1.}
    \label{fig:kernel_fusion}
  \end{minipage}
\end{figure}

Programs for accelerators are typically decomposed into a series of smaller subprograms, known as \textit{kernels}, which are executed atomically, and possibly concurrently, in some order on the hardware. Each subprogram can consist of many elementary operations and can simultaneously use all subsystems. For example, for a matrix multiplication followed by bias addition and activation, the entire computation can be carried out in a single
\textit{fused} kernel \citep{snider2023operatorfusionxlaanalysis, cloud_tpu_perf_guide}, where matrix unit outputs are immediately processed by scalar/vector units (for bias addition and activation) before writing them to memory. This minimizes the overhead associated with launching and terminating kernels, avoids multiple round-trips to the memory, and enables more effective simultaneous use of the different compute resources. Figure \ref{fig:kernel_fusion} illustrates this concept.

\subsection{Ridge point analysis}
\label{sec:runtime_perf_analysis}

Depending on the kernel, different subsystems of an accelerator are utilized to varying degrees, with one often becoming the bottleneck that dictates the kernel's overall runtime. To accurately model performance and identify the bottleneck, we must quantify the capabilities of each accelerator subsystem and how the kernel uses it \citep{chern2022tpuknnknearestneighbor}.

We quantitatively characterize an accelerator's performance by measuring the peak throughput of each of its subsystems. For example, we measure the peak memory bandwidth for each memory subsystem and the peak operations per second for each compute subsystem. Depending on the subsystems used and their extent of saturation, we may choose to model only the relevant subsystems. For example, if a subsystem is not used or contributes negligibly to the runtime, it can be omitted.

For ease of exposition, we present the analysis for TPUs and focus on three key subsystems: (i) HBM memory, (ii) the VPU and (iii) the MXU. However, this analysis applies to GPUs too, with MXU mapped to TensorCores and VPU to CUDA cores. We define the following parameters to quantify the throughput of each subsystem:
\begin{itemize}
    \item $\beta$: maximum HBM bandwidth in bytes per second.
    \item $\gamma$: maximum number of VPU operations per second.
    \item $\pi$: maximum number of MXU operations per second.
\end{itemize}

Similarly, we characterize a kernel by measuring its utilization of each subsystem in its lifetime.
\begin{itemize}
    \item $M$: number of bytes transferred to/from the HBM.
    \item $O_\text{VPU}$: number of operations executed on the VPU.
    \item $O_\text{MXU}$: number of operations executed on the MXU.
\end{itemize}

Since the kernel can utilize all subsystems simultaneously, its total runtime is determined by the subsystem that requires the most time to complete its work.\footnote{In some cases, there may be dependencies between subsystems that may cause the dominating subsystem to stall. However, in practice, most optimized kernels do not suffer significantly from such dependencies and saturate at least one of the subsystems.} Therefore, we can estimate the total runtime of the kernel as:
\begin{equation}
\text{runtime} = \max\left(\frac{M}{\beta}, \frac{O_\text{VPU}}{\gamma}, \frac{O_\text{MXU}}{\pi}\right). \label{eqn:runtime}
\end{equation}

The bottleneck is the subsystem corresponding to the largest argument to $\max(\dots)$ in \eqref{eqn:runtime}. For example, a kernel is considered to be \textit{memory-bound} if its memory subsystem cannot feed data sufficiently fast to keep compute subsystems busy, i.e., $\frac{M}{\beta} \ge \max(\frac{O_\text{VPU}}{\gamma}, \frac{O_\text{MXU}}{\pi})$. To minimize overall runtime, we must address the bottleneck subsystem until it is no longer the limiting factor.

A corollary of this model is that increasing the utilization of non-bottleneck subsystems may \textit{not} necessarily increase the kernel's runtime. To formalize this, we define \textit{ridge points} of \eqref{eqn:runtime}, which are configurations where the runtimes of any two subsystems are equal. For example, we can estimate as $\frac{\gamma}{\pi}$ the maximum number of VPU operations that can be performed per MXU operation to remain MXU-bound. Since the MXU has much higher throughput, i.e., $\pi \gg \gamma$, it is often more convenient to use the number of VPU operations per d-dimensional dot product on the MXU, i.e., $\frac{\gamma}{\left(\frac{\pi}{2d}\right)}$. This reformulation helps keep the ratio a small and interpretable integer. We can similarly define quantities such as $\frac{\gamma}{\left(\frac{\beta}{4}\right)}$ to denote the number of VPU operations that can be performed per 4-bytes of data transferred to/from the HBM. Ridge points thus provide a simple way to understand the balance between different subsystems and easily reason about how changes in a subsystem's utilization can impact overall performance \citep{10.1145/1498765.1498785}. For a list of values of these quantities on different accelerators, see Table \ref{tbl:hw_throughput}.

\begin{table*}[t]
\caption{\textbf{Peak throughput and ridge points of different subsystems in accelerators.} The $\gamma$ for TPUv4 was taken from \citet{chern2022tpuknnknearestneighbor}, and the $\gamma$ for TPUv5e was estimated by timing VPU-bound kernels (see Appendix \ref{ssec:reverse_engineer_vpu_flops}). All other values derived directly from the official datasheets.}
\label{tbl:hw_throughput}
\vskip 0.15in
\begin{center}
\begin{small}
\begin{sc}
\begin{tabular}{cccccc}
\toprule
Device & $\beta$ & $\gamma$ (TFLOP/s) & $\pi$ (TFLOP/s) & $\frac{\gamma}{\left(\frac{\pi}{256}\right)}$ & $\frac{\gamma}{\left(\frac{\beta}{4}\right)}$\vspace{2pt} \\
& & fp32 & bf16 & ops per 128-d dot & ops per 4 bytes \\
\midrule
A100 PCIe & 1.935 TB/s & 19.5 & 312 & $\approx 16$ & $\approx 40$ \\
H100 SXM & 3.35 TB/s & 67 & 1,979 & $\approx 8$ & $\approx 80$ \\
TPUv4   & 1.2 TB/s & 4.3 & 275 & $\approx 4$ & $\approx 14$  \\
TPUv5e   & 819 GB/s & $\approx 6.14$ & 197 & $\approx 8$ & $\approx 30$ \\
\bottomrule
\end{tabular}
\end{sc}
\end{small}
\end{center}
\vskip -0.1in
\end{table*}

\section{Problem Setup} \label{sec:problem}

Given a matrix $W\in \R^{n \times d}$ and a vector $x \in \R^d$, the task is to approximately find the $K$ largest elements of $y := Wx \in \R^n$.

\textbf{Expected recall.} Let $U$ represent the set of actual top-K elements of $y$, and let $V$ represent the top-K elements returned by an approximate algorithm. We define \textit{expected recall} as the expected fraction of the true top-K elements retrieved by the algorithm, assuming that the top-K elements are placed randomly and uniformly in $y$:

\vskip -0.05in
\[
\Ex{\text{recall}} = \Ex{\frac{|U \cap V|}{|U|}}.
\]

\textbf{Objective.} The goal is to \textit{minimize the time} required for this operation while \textit{maximizing expected recall}, thereby improving the Pareto frontier describing the trade-off between latency and expected recall objectives.

\section{Related Work}
\label{sec:related_work}

As of this writing, \texttt{jax.lax.top\_k} is the only exact Top-K implementation available for TPUs. \citet{chern2022tpuknnknearestneighbor} proposed a faster approximate Top-K algorithm, which is currently exposed in JAX as \texttt{jax.lax.approx\_max\_k}. To the best of our knowledge, these are the only Top-K implementations available for TPUs.

A large body of prior work focuses on exact Top-K algorithms for GPUs \citep{Xie2024RTopKUR, Gaihre2021DrTD, Zhang2023ParallelTA, Li2024RadiKSA, Dashti2013EfficientCO, Alabi2012FastA, Shanbhag2018EfficientTQ, 8778227}. Since the algorithmic framework introduced by \citet{chern2022tpuknnknearestneighbor}, which we generalize in this work, can transform any exact algorithm into a faster approximate variant, we do not compare with exact algorithms. We provide a detailed description of the \citet{chern2022tpuknnknearestneighbor} algorithm in Section \ref{sec:original_algorithm}.

\citeauthor{key2024approximate} independently develope the same generalized algorithm, implement it for GPUs in the unfused setting, and evaluate quality in several downstream tasks including SparQ attention \cite{ribar2024sparqattentionbandwidthefficientllm} and link-prediction in knowledge graphs. Although the algorithm is shared, our work provides several distinct theoretical and practical advances.  Our probabilistic model of the algorithm is exact, whereas theirs relies on approximations that lead to noticeable differences (see Appendix Section \ref{asec:concurrent_work}). They provide an API, \texttt{approx\_top\_k(array, K, K', B)}, that requires manual tuning of algorithm parameters. In contrast, with our theoretical analysis along with detailed runtime performance modeling, we provide a fast automatic parameter selection routine and a more user-friendly interface, \texttt{approx\_top\_k(array, K, recall\_target)}. Beyond the unfused setting, we also implement a matmul-fused implementation that extracts non-trivial speedups in realistic workloads. The ridge-point analysis in Section 2.3, combined with our analysis of the implementation in Section 6.3, provides a principled framework for determining the available headroom in a fusion and enables fusion-aware parameter selection, which could, in principle, be adapted into a cost model to support automatic compiler-driven fusion decisions.


\section{The Original Algorithm}
\label{sec:original_algorithm}

\citet{chern2022tpuknnknearestneighbor} designed an approximate Top-$K$ algorithm that operates in two stages. In the first stage, the input array is partitioned by grouping elements separated by a fixed stride into buckets. The top-1 element of each bucket is gathered to form a \textit{smaller} array. In the second stage, this array is sorted using bitonic sort, and the top K elements are returned. The first stage reduces the size of the input array for the \textit{expensive} second stage, which improves performance. Approximations occur when multiple top-K elements fall into the same bucket since only one is selected and the rest are discarded. Increasing the number of buckets reduces the likelihood of such collisions and can be chosen to achieve a desired average recall target. Appendix Figure \ref{fig:top1algo} presents the algorithm.

The algorithm accepts \texttt{recall\_target} as a parameter, denoted by $r$, which specifies the desired expected recall. The required number of buckets, denoted by $B$, is calculated using a closed-form expression that relates $B$ to the expected recall under a model in which the positions of the top-K elements are independently and uniformly distributed over the input array:

\[
 B \geq \frac{1}{1 - \mathop{\mathbb{E}}[\text{Recall}]^{\frac{1}{K - 1}}}  \approx \frac{K - 1}{1 - r}.
\]

The input often results from a matrix multiplication, e.g., a maximum inner-product search (MIPS) or the Top-$K$ on key-query logits in attention. The algorithm's first stage, which executes on the scalar/vector units, can often be fused with the preceding matrix multiplication, which executes on the matrix units. Hence, the fused first stage might incur little to no additional cost since it utilizes the otherwise idle scalar/vector compute units while the matrix units are busy.

To design their algorithm and fused implementation, the authors adopt a principled approach by modeling accelerator performance, similar to the model described in Section \ref{sec:runtime_perf_analysis}. Their first stage uses a fixed budget of three operations per element to track the top-1 element (and its index) of each bucket. We argue that this leaves compute resources underutilized in many cases, as detailed below:
\begin{enumerate}
    \item Their analysis focuses on matrix multiplications with 128-dimensional dot products, which on most hardware provide only 4-8 vector operations per output element, i.e., each 128-d dot product, consistent with their budget of three operations per element. However, we frequently work with larger dimensions, where the available scalar/vector compute per output element is nearly $\frac{\text{dims}}{128}$ times higher than the numbers they estimate.
    \item Even for 128-dimensional dot products, the first stage may not saturate the scalar/vector units on all hardware platforms. See Table \ref{tbl:hw_throughput}.
    \item In memory-bound computations, more scalar/vector compute is available than would be expected from a matrix-multiplication-bound computation.
\end{enumerate}

The additional compute available enables more sophisticated algorithms for the first stage, potentially improving the recall with fewer elements to process in the second stage. A more expensive first stage might still yield overall gains if gains in the second stage outweigh the increased cost of the first stage. Based on these insights, we generalize their algorithm to more \textit{flexibly} utilize the available compute by selecting \textit{the top-$K'$ elements from each bucket instead of just the top-$1$.}

\section{Method}
\label{sec:method}

We describe our algorithm in Section \ref{sec:methods:algorithm} and provide an analysis in Section \ref{sec:methods:analysis}. In Section \ref{sec:methods:implementation}, we discuss the key ideas in our implementation of the algorithm for TPUs.

\subsection{Algorithm}
\label{sec:methods:algorithm}


\begin{tabular}{p{0.48\linewidth} p{0.48\linewidth}}
\textbf{Step 1: Partition into buckets} & \textbf{Step 2: Select Top-$K'$ per bucket} \\[1mm]
Given \(A = [a_1, \dots, a_N]\), partition into \(B\) buckets:  
\[
G_i = \{ a_{i + jB} \mid j \in \mathbb{Z}_{\ge 0}, i + jB \le N \}, \quad i = 1, \dots, B
\]
&
Select the top-$K'$ elements from each bucket:  
\[
T_{K'}(G_i) = \text{Top-K'}(G_i)
\] \\[2mm]

\textbf{Step 3: Merge selections} & \textbf{Step 4: Return approximate top-$K$} \\[1mm]
Merge the top-$K'$ elements from all buckets:  
\vspace{-6pt}
\[
A_\text{selected} = \bigcup_{i=1}^{B} T_{K'}(G_i)
\]
&
Sort the merged set and return the top-$K$ elements:  
\[
T_K^\text{approx}(A) = \text{sorted}(A_\text{selected})[:K]
\]
\end{tabular}






\subsection{Analysis}
\label{sec:methods:analysis}

Consider a scenario in which we have $N$ balls, $K$ of which are special balls, and $B$ buckets. The $N$ balls are evenly distributed in the $B$ buckets. To model the distribution process, we can randomly order all the balls and then partition them into buckets: the first $\frac{N}{B}$ balls go to the first bucket, the next $\frac{N}{B}$ balls go to the second bucket, and so on. In the context of our algorithm, the $N$ balls correspond to the input elements, the $K$ special balls represent the top-K elements, and $B$ buckets correspond to the ``buckets" in the algorithm.

Let $X_b$ be a random variable that denotes the number of special balls in bucket $b$. Approximation errors occur when more than $K'$ special balls are placed in the same bucket. The total number of excess collisions is given by the sum of excess special balls in each bucket.

\begin{align*}
\Ex{\text{Excess-collisions}} &= \Ex{\sum_{b=1}^B \max(0, X_b - K')}\\
&= \sum_{b=1}^B \Ex{\max(0, X_b - K')}.\\
\end{align*}

There exists a joint probability distribution that governs the set of $X_b$ that satisfies the constraint that the total number of special balls in all buckets sums to $K$, i.e., $\sum_{b=1}^B X_b = K$. However, the marginals $X_b$ are all identically distributed as $X_b \sim \text{Hypergeometric}(N, K, \frac{N}{B})$. This arises because the distribution of special balls in the first bucket must be the same as in all other buckets by symmetry, and it is apparent that the distribution of special balls in the first bucket must follow $\text{Hypergeometric}(N, K, \frac{N}{B})$. This is sufficient to simplify further:

\[ \Ex{\text{Excess-collisions}} = B \times \Ex{\max(0, X_0 - K')}. \]

The number of excess collisions is related to the recall as follows.

$$\Ex{\text{Recall}} = 1 - \frac{\mathop{\mathbb{E}}[\text{Excess-collisions}]}{K}.$$

In Appendix \ref{ssec:mc_expr_verification}, we verify the accuracy of Monte Carlo evaluations of this expectation against the recall obtained from the simulated runs of the algorithm. Theorem \ref{thm:exp_recall} derives an algebraic expression for this expectation.

\citet{chern2022tpuknnknearestneighbor} model their algorithm as randomly distributing $K$ balls in $B$ buckets and relate it to the classical birthday problem. Based on this model, they derive a bound on the expected recall and invert the expression to obtain a formula for the number of buckets. However, their analysis neglects several structural constraints of the problem: (i) the number of balls in each bucket cannot exceed $\frac{N}{B}$, (ii) the balls are sampled without replacement, and (iii) only the non-colliding balls are counted as correctly retrieved, even though \textit{one} of the colliding balls is always correctly retrieved in a bucket with collisions. \textit{In Theorem \ref{thm:exp_recall}, we derive a new bound on the number of buckets for $K'=1$ based on our model that is provably tighter than theirs by at least a factor of two.} We verify the quality of our bounds in Appendix \ref{ssec:k1_bound_verification} and show that it closely approximates the exact expression with high fidelity.

\newtheorem{theorem}{Theorem}
\begin{theorem} 
\label{thm:exp_recall}
Suppose $N$ balls are randomly distributed into $B$ buckets $G_1, \cdots, G_B$, each getting $N/B$ balls. The recall of the top-$K'$ balls across all the $B$ buckets with respect to the top-$K$ balls overall is given by:
$\mathop{\mathbb{E}}[\textrm{Recall}] = 1 - \frac{B}{K} \times \sum_{r=K'+1}^{\min(K,N/B)} (r-K') \frac{{K \choose r} {{N-K} \choose {\frac{N}{B}-r}}}{{N \choose {\frac{N}{B}}}}$.
Specifically, for $K'=1$ and a target recall factor of at least $r$, the bound below implies that $B = \frac{K}{2 \left(1- r + \frac{K}{2N}\right)}$ suffices to guarantee the target recall $r$.
\end{theorem} 
\textbf{Remark}. Note that for $K'=1$, the bound is a factor of $2$ tighter than that in~\citet{chern2022tpuknnknearestneighbor}.

We provide the proof in Appendix \ref{asec:th1_proof}.

\subsection{Implementation}
\label{sec:methods:implementation}

In the first stage, we take an input array of shape \texttt{[batch\_size, N]} and output two vectors: one for values and another for indices, both of which have the shape \texttt{[batch\_size, B $\times$ K']}. Here, $N$ is the total number of elements, $B$ is the number of buckets, and $K'$ is the number of top elements we select from each bucket.
  
We focus on identifying the top-$K'$ elements of a single bucket since supporting multiple buckets is a matter of running many \textit{independent} instances of this subroutine.
To create an effective fusible implementation, we track the top-$K'$ elements in an online fashion as inputs continuously stream in from the matrix multiplication unit. We maintain two lists per bucket, one for the top-K' \texttt{values} and another for their corresponding \texttt{indices}. The \texttt{values} list is kept in descending order, and we ensure that each value's corresponding index is at the same position in the \texttt{indices} list. When a new element arrives, we update the lists in two steps:
\begin{enumerate}
    \item If the new element is larger than the smallest element in the \texttt{values} list, we replace the smallest element (and its index) with the new element (and its index).
    \item We perform a single bubble sort pass over the lists to move the new element to its correct position.
\end{enumerate}

\NewDocumentCommand{\ARef}{ s s m }{%
    \IfBooleanTF{#2}{}{%
        \cref{#3}%
    }%
    \IfBooleanT{#1}{%
        \IfBooleanF{#2}{%
            , %
        }%
        Line~\ref{#3}%
    }%
}

Algorithm \ref{alg:sorted_topk_insertion} contains the pseudocode for this subroutine. The first step requires one comparison and two selects for updating the value and index. The second step requires comparing adjacent elements (one comparison) and conditionally swapping elements (four selects) for each of the $(K' - 1)$ positions. In total, each input element requires $(5K' - 2)$ operations.

Since the \texttt{values} list is stored in descending order, an input element larger than the k-th value in the list is also larger than all subsequent values. This property allows the comparison in~\ARef**{alg:sorted_topk_insertion:loop_cmp}, Algorithm~\ref{alg:sorted_topk_insertion} to be done using the input element as the LHS, which eliminates a loop-carried dependency.

Note that Algorithm~\ref{alg:sorted_topk_insertion} does not include an early return if the condition on~\ARef**{alg:sorted_topk_insertion:if_cmp} fails, nor does it exit the bubble sort loop early when the condition in~\ARef**{alg:sorted_topk_insertion:loop_cmp} fails. This is required to vectorize the routine across buckets. An early return would theoretically reduce operations for individual buckets but would prevent vectorization across buckets.

Once all inputs are processed, we obtain the final result by separately merging all \texttt{values} lists to obtain the first-stage values list and merging the corresponding \texttt{indices} lists to obtain the first-stage indices list.

\renewcommand\algorithmiccomment[1]{%
  \textcolor{olive}{\hfill\ \eqparbox{COMMENT}{/* #1 */}}%
}

\begin{figure}[tbp]
\begin{minipage}{0.53\linewidth}
\begin{algorithm}[H]
    \caption{\texttt{TopKPrimeUpdate}}
    \label{alg:sorted_topk_insertion}
\begin{algorithmic}[1]
    \STATE \textbf{Input:} input, index, values[K'], indices[K']
    \STATE \textbf{Output:} values[K'], indices[K']
    \STATE \textbf{Precondition:} values sorted in descending order
    \IF[one compare]{input $\ge$ values[K']}
        \label{alg:sorted_topk_insertion:if_cmp}
        \STATE values[K'] = input \COMMENT{one select}
        \STATE indices[K'] = index \COMMENT{one select}
    \ENDIF
    \FOR{$k=K'$ \textbf{to} $2$}
        \IF[one compare]{values[k] $>$ values[k-1]} \label{alg:sorted_topk_insertion:loop_cmp}
           \STATE swap(values[k], values[k-1]) \COMMENT{two selects}
           \STATE swap(indices[k], indices[k-1]) \COMMENT{two selects}
        \ENDIF
    \ENDFOR
\end{algorithmic}
\end{algorithm}
\end{minipage}
\hfill
\begin{minipage}{0.47\linewidth}
\begin{algorithm}[H]
    \caption{Vectorized Top-$K'$ Update}
    \label{alg:vectorized_body}
\begin{algorithmic}[1]
    \STATE \textbf{Input:} input array of size $N$, number of lanes $L$, number of buckets $B$, top-$K'$ values (-$\infty$ initialized) and indices lists of shape [B, K']
    \STATE \textbf{Output:} top-$K'$ values and indices lists
    \STATE $num\_chunks \gets N / L$
    \FOR{$in\_chunk \gets 0$ \textbf{to} $num\_chunks - 1$}
        \STATE $out\_chunk \gets in\_chunk \bmod (B / L)$
        \STATE \textbf{Load} inputs corresponding to $in\_chunk$
        \STATE \textbf{Load} current top-$K'$ lists for $out\_chunk$
        \STATE \textbf{Compute} updated top-$K'$ lists using trivially vectorized \texttt{TopKPrimeUpdate}
        \STATE \textbf{Store} updated top-$K'$ list at $out\_chunk$
    \ENDFOR
\end{algorithmic}
\end{algorithm}
\end{minipage}
\end{figure}


Since buckets group elements separated by a fixed stride, contiguous input elements map to different buckets. We can logically view the input array as having  the shape \texttt{[batch\_size, N / B, B]}. We store the top-K' lists with a physical layout of \texttt{[batch\_size, K', B]} so that the minor-most axis maps to the bucket axis, which aligns with the input's logical shape. The top-K' update subroutine (Algorithm \ref{alg:sorted_topk_insertion}) can be executed independently for each bucket and is trivially vectorizable along the bucket axis. 

To simplify the implementation, we restrict the number of buckets to a multiple of the vector width, denoted by $L$. We process contiguous L-sized chunks of the inputs and their corresponding L-sized top-K' lists in each iteration of a vectorized loop. Algorithm \ref{alg:vectorized_body} outlines the vectorized version. Although this implementation appears to require $2K'$ loads and stores of the top-K' lists for each input that is read, we can schedule the loop iterations so that the input chunks corresponding to the same bucket are executed consecutively; this lets the top-K' lists fully reside in the registers or the nearest cache depending on the choice of $K'$ and the hardware.

Based on these insights, we implement our first-stage kernel in Pallas, a JAX kernel language~\citep{jax2018github}. We use \texttt{jax.lax.sort\_key\_val} and slice the top-K elements for the second stage. We share the Python code for our algorithm with detailed comments in Listing~\ref{lst:pallas_implementation} for the unfused implementation and Listing~\ref{lst:fused_pallas_implementation} for the matmul-fused implementation. To find the algorithm parameters for a given input shape and recall target, we sweep through legal configurations and list those that meet the recall target. We then heuristically choose the configuration with the best performance. To calculate expected recall, we use Monte Carlo evaluations of the expectation expression derived in Section \ref{sec:methods:analysis}. We share the Python code to estimate expected recall and select algorithm parameters in Listing \ref{lst:parameter_selection}. The computational cost of this procedure is relatively low; see Appendix Section \ref{ssec:param_sweep} for a detailed discussion.


\section{Results and Discussion}
\label{sec:results}

In Section \ref{sec:results_accuracy}, we show that our algorithm substantially reduces the input size for the same expected recall compared to the improved baseline, which is the original algorithm by \citet{chern2022tpuknnknearestneighbor} with our improved bound. Section \ref{sec:results_latency} discusses the performance of our unfused Pallas implementation for TPUs, and Section \ref{sec:results_fused_latency} does the same for our matmul-fused implementation.

\subsection{Theoretical effectiveness of the first-stage filtering}
\label{sec:results_accuracy}

Table \ref{tbl:results_top1024_262144} shows the relationship between $K'$, the number of buckets, and the expected recall to select the top-1024 elements of an array of 262,144 elements. With a fixed number of output elements ($K' \times num\_buckets$), the expected recall increases significantly with $K'$. Keeping the expected recall fixed, even small values of $K'$ ($\leq 4$) substantially reduce the number of output elements. For example, to achieve an expected recall of 95\%, $K'=1$ requires at least 16,384 output elements, while $K'=4$ requires only 2,048 elements, an 8$\times$ reduction in the number of elements to process. 

Appendix Figure~\ref{fig:accuracy_plot} plots expected recall versus the number of output elements for different values of $K'$ to select the top-3,360 elements from an array of 430,080 elements. The expected recall improves rapidly with increasing $K'$, as highlighted by the clear separation between the curves corresponding to our algorithm and the baseline.

\begin{table*}
\caption{\textbf{(Left) Expected recall versus K' for selecting top-1024 elements from an array of 262,144 elements.} \texttt{\#num\_elements} refers to the number of output elements from the first stage, which is $B \times K'$. A smaller \texttt{\#num\_elements} leads to better performance in the second stage. \textbf{(Right) The runtime of our algorithm on TPUv5e for a batch size of 8.} The \texttt{jax.lax.approx\_max\_k} rows present the performance of the official JAX implementation 
(which supports only $K'=1$), while the $K'=1$ rows present the performance of our implementation.}
\label{tbl:results_top1024_262144}
\vskip 0.15in
\begin{center}
\begin{small}
\begin{sc}
\begin{tabular}{lc|cc|cccr}
\toprule
\multicolumn{2}{c|}{Algorithm Parameters} & \multicolumn{2}{c|}{Algorithmic Performance} & \multicolumn{3}{c}{Runtime Performance} \\
K' & \texttt{buckets} & \texttt{num\_elements} & $\Ex{\text{recall}}$ & Stage 1 & Stage 2 & Total \\
\midrule
\scriptsize{\texttt{jax.lax.approx\_max\_k}} & 131,072 & 131,072 & 0.998 $\pm$ 0.000 & 12us & 649us & 661us \\
\scriptsize{\texttt{jax.lax.approx\_max\_k}} & 65,536 & 65,536 & 0.992 $\pm$ 0.001 & 13us & 292us & 305us \\
\scriptsize{\texttt{jax.lax.approx\_max\_k}} & 32,768 & 32,768 & 0.987 $\pm$ 0.004 & 13us & 131us & 144us \\
\midrule
1    & 65,536 & 65,536 & 0.992 $\pm$ 0.001 & 13us  & 313us & 326us \\
1    & 32,768 & 32,768 & 0.987 $\pm$ 0.004 & 14us & 141us & 155us \\
1    & 16,384 & 16,384 & 0.972 $\pm$ 0.005 & 13us & 64us & 77us \\
1    & 8,192 & 8,192 & 0.942 $\pm$ 0.007 & 13us & 30us & 42us \\
2    & 4,096 & 8,192 & 0.991 $\pm$ 0.003 & 15us & 30us & 45us \\
2    & 2,048 & 4,096 & 0.968 $\pm$ 0.006 & 13us & 14us & 27us \\
3    & 2,048 & 6,144 & 0.996 $\pm$ 0.002 & 16us & 32us & 48us \\
3    & 1,024 & 3,072 & 0.977 $\pm$ 0.005 & 12us & 11us & 23us  \\
4    & 1,024 & 4,096 & 0.996 $\pm$ 0.002 & 13us & 14us & 27us \\
4    & 512 & 2,048 & 0.963 $\pm$ 0.007 & 12us & 8us & 20us \\
5    & 512 & 2,560 & 0.989 $\pm$ 0.004 & 13us & 9us & 22us \\
6    & 512 & 3,072 & 0.997 $\pm$ 0.002 & 14us & 11us & 25us \\
6    & 256 & 1,536 & 0.951 $\pm$ 0.008 & 14us & 8us & 22us \\
8    & 512 & 4,096 & 0.992 $\pm$ 0.004 & 16us & 14us & 30us \\
10   & 256 & 2,560 & 0.999 $\pm$ 0.000 & 19us & 9us & 28us \\
12   & 128 & 1,536 & 0.984 $\pm$ 0.006 & 23us & 8us & 31us \\
16   & 128 & 2,048 & 0.999 $\pm$ 0.001 & 29us & 8us & 37us \\
\bottomrule
\end{tabular}
\end{sc}
\end{small}
\end{center}
\vskip -0.1in
\end{table*}

Figure \ref{fig:factor_heatmap} shows the factor by which our algorithm variants, up to $K'=4$, reduce the size of the second-stage input across a broad range of configurations ({$\frac{K}{N} \in \{0.01\%, \ldots, 25\% \}$ and array sizes $\in [256, 4e9]$) over the baseline $K'=1$. We account for the implementation constraints necessary for simplicity and performance such as requiring the number of buckets to be a multiple of 128 and divide the input array size evenly; therefore, the numbers indicate \textit{real realizable reductions} using our implementation. The figure demonstrates that our algorithm significantly reduces the number of elements in virtually all configurations, with a median reduction of 7$\times$. In particular, the $K' > 1$ variants perform worse or equal to $K'=1$ only for very small values of K ($K \le 10$), which is a consequence of our implementation constraints where a smaller $K'$ can sometimes yield a better-aligned $B$ that results in fewer total $B \cdot K'$ elements after rounding. However, since we always select the best $K'$ in $[1, 4]$, our method never performs worse than the baseline by construction. We conclude that our algorithm is broadly applicable and effectively reduces the number of output elements. 

\begin{figure}[h]
\includegraphics[width=\linewidth]{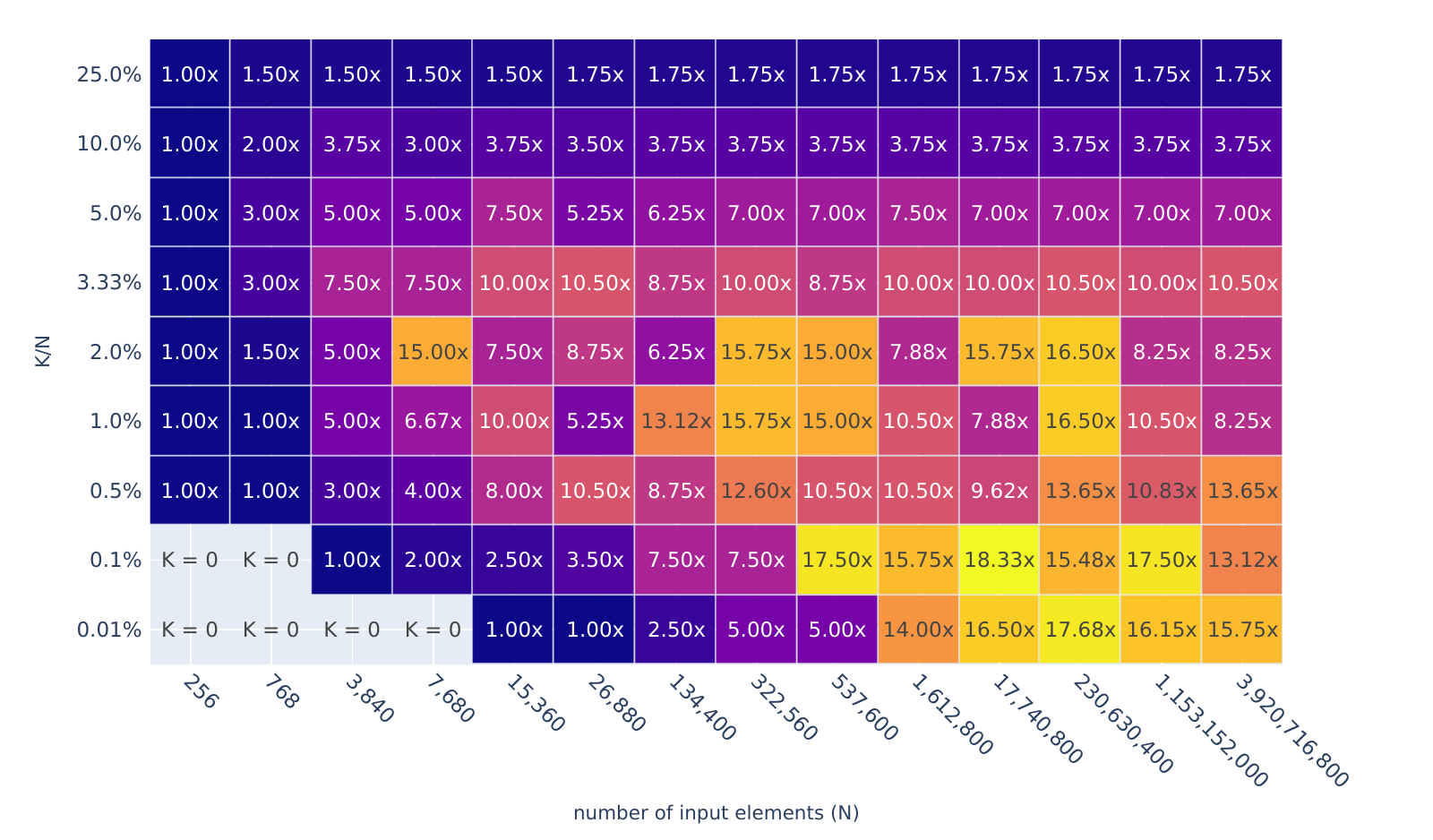}
\caption{\textbf{Factor of reduction in output elements over the baseline ($K'=1$) for 99\% expected recall target.} The heatmap shows the factor by which our generalized algorithm with $1 \le K' \le 4$ reduces the number of elements in the first stage over the reductions provided by the baseline, i.e., a value of 2$\times$ indicates that our algorithm outputs two times fewer elements compared to the $K'=1$ baseline. Our implementation constrains the number of buckets to be a multiple of 128 and divide the input size for simplicity and performance, which is accounted for in this figure. Even though $K' > 1$ would require fewer buckets compared to $K' = 1$, rounding the number of buckets to satisfy the constraints may lead to more output elements than required by $K'=1$.}
\label{fig:factor_heatmap}
\end{figure}

\subsection{Improved latency of finding the Top-K elements on TPUv5e}
\label{sec:results_latency}

\DeclareSIUnit{\microsecond}{\SIUnitSymbolMicro s}

Table \ref{tbl:results_top1024_262144} presents the latency of our unfused implementation to identify the top-1024 elements from an array of 262,144 elements. To achieve a recall target of 99\%, the baseline requires 305\si{\microsecond}, while our algorithm with $K'=4$ takes only 27\si{\microsecond}, resulting in an 11$\times$ speedup.

The first stage cost remains nearly constant from $K'=1$ to $K'=6$ due to its memory-bound nature. According to the performance model in Section \ref{sec:runtime_perf_analysis} and Table \ref{tbl:hw_throughput}, the first stage must be memory bound until we exceed 30 VPU operations per 4-byte element, which occurs around $K'=6$, according to the operation count formula in Section \ref{sec:methods:implementation}. Therefore, we expect the latency of the first stage to be independent of $K'$ until we reach $K'=6$. 

\subsection{Fusing Top-$K$ with matrix multiplication}
\label{sec:results_fused_latency}

Many real-world applications require identifying the Top-$K$ results from the outputs of matrix multiplications. One prominent example is maximum inner-product search (MIPS), where, for a given query vector, the task is to retrieve the top-K vectors from a large database that have the highest inner products.

\begin{table}
\caption{\textbf{The runtime of our algorithm on TPUv5e to identify the top-1024 elements from a database of 1M 128-dimensional vectors with 99\% recall for 1024 query vectors.} The \texttt{jax.lax.top\_k} row represents the performance of exact Top-K. The \texttt{jax.lax.approx\_max\_k} row presents the performance of the official JAX implementation for the $K'=1$ setting. The remaining rows present the performance of our implementation.}\label{tbl:mips_latency}
\vskip 0.15in
\begin{center}
\begin{small}
\begin{sc}
\begin{tabular}{ccccc}
\toprule
Algorithm & Matmul & Stage 1 & Stage 2 & Total \\
\midrule
\scriptsize{\texttt{jax.lax.top\_k}} & 7.32ms & - & 587ms & 594ms \\
\scriptsize{\texttt{jax.lax.approx\_max\_k}} & 9.06ms & fused & 118ms & 127ms \\
$K'=1$ & 7.32ms & 6.58ms & 50.0ms & 64ms \\
$K'=1$ & 9.03ms & fused & 50.0ms & 59ms \\
$K'=4$ & 7.31ms & 10.80ms & 3.51ms & 22ms \\
$K'=4$ & 6.55ms & fused & 3.51ms & 10ms \\
\bottomrule
\end{tabular}
\end{sc}
\end{small}
\end{center}
\vskip -0.1in
\end{table}

We evaluate our algorithm on a MIPS workload of one million 128-dimensional vectors and 1024 queries. Table \ref{tbl:mips_latency} reports runtimes on TPUv5e. Exact Top-$K$ (\texttt{jax.lax.top\_k}) takes 80$\times$ longer in the second stage than the matmul (587ms vs 7.32ms). \texttt{jax.lax.approx\_max\_k} reduces this to 13$\times$ (118ms), and our $K'=1$ unfused implementation to 6$\times$ (50ms).

Moving to $K'=4$, the second stage (3.51ms) falls below half the matmul cost, leaving the matmul (7.31ms) and first stage (10.80ms) as the bottleneck. Fusing the first stage with the matmul eliminates its cost and improves matmul performance (6.55ms). The gains from fusion can be significant in practice. The MIPS matmul multiplies \texttt{[B, D]} by \texttt{[D, N]} (queries, vector size, database size), giving arithmetic intensity $\frac{2}{E}\min(B,D)$ (derived in Appendix~\ref{sec:aimatmul_derive}), where $E$ is the element size. In large-scale deployments with $D$ in the low hundreds, the matmul is often memory-bound. Fusion avoids writing the large output tensor to memory, increasing arithmetic intensity and shifting the matmul closer to compute-bound.


 \section{Conclusion}

We present a generalization of the approximate Top-$K$ algorithm of \citet{chern2022tpuknnknearestneighbor} that selects the top-$K'$ elements per bucket in the first stage, instead of restricting to top-1. We motivate this generalization through a principled analysis of the runtime behavior of the original algorithm, showing that the choice of $K'=1$ does not fully utilize the available compute resources. This observation naturally leads to a broader family of algorithms parameterized by $K'$, which can substantially reduce the second-stage workload while maintaining high recall. We theoretically analyze the generalized algorithm, deriving an exact expression for the expected recall, and obtain a bound for $K’=1$ which is provably tighter by a factor of 2 compared to that in \citet{chern2022tpuknnknearestneighbor}. For $K'>1$, we demonstrate both theoretically and empirically that the algorithm substantially reduces the second-stage input size across a wide range of configurations, spanning $\frac{K}{N}$ ratios from 0.01\% to 25\%, with a median reduction of $7\times$ over the $K'=1$ baseline.

To realize these gains in practice, we implement the algorithm for Cloud TPUv5e together with a principled performance modeling framework that accurately predicts runtimes and enables a user-friendly \texttt{approx\_top\_k(array, K, recall\_target)} interface that requires no manual tuning. We additionally present a matmul-fused implementation that further eliminates the first-stage cost. While our kernel implementations are specific to TPUs, the algorithm, performance modeling and its analysis are general, and we believe that it has applicability to other hardware platforms such as GPUs.

\textbf{Limitations and Future Work.} Our empirical evaluation is focused on Cloud TPUv5e, and extending empirical evaluation and the matmul-fused implementation to GPUs is an important open direction. The recall analysis assumes uniform random placement of top-$K$ elements; efficient parallel randomization schemes that bring real-world inputs in line with this model would fully close the gap between theory and practice. While we derive a provably tighter bound for the $K'=1$ setting, extending similarly tight numerically stable analytical bounds to the general $K'>1$ setting remains open. In particular, deriving closed-form approximations that eliminate the need for Monte Carlo estimation during parameter selection would improve the practicality of the approach. A further limitation is that our analysis focuses on expected recall and does not characterize its variance or the full error distribution. In particular, understanding the variability induced by collisions could yield a more complete picture of the algorithm. Integrating the algorithm into an optimizing compiler would enable automatic fusion decisions and fusion-aware parameter selection. Our generalization of the first stage represents the simplest step in a richer design space; more sophisticated first-stage selection schemes beyond the top-$K'$ per bucket approach could push the performance further.

\clearpage

\bibliography{main}
\bibliographystyle{tmlr}

\clearpage
\appendix
\section{Appendix}

\subsection{Estimating peak VPU throughput of TPUv5e.}
\label{ssec:reverse_engineer_vpu_flops}

We used two test programs with a controllable parameter that allows us to vary the number of vector operations per element. We run the programs on a large \texttt{s32[4096, 4096]}-shaped array with different parameters and time the kernels. We verify that the compiler fuses the operations into a single kernel. We assume that addition and multiplication are instructions in the TPU's ISA. Given the large size of the inputs, we assume that these programs saturate the vector processing unit. 

\begin{lstlisting}[language=Python]
@partial(jax.jit, static_argnames='n')
def fibonacci(x, y, n):
  for i in range(n):
    c = x + y
    x = y
    y = c
  # We expect the compiler to optimize the snippet to the following sequence:
  # r1 = x
  # r2 = y
  # r3 = r1 + r2
  # r1 = r2 + r3
  # r2 = r3 + r1
  # r3 = r1 + r2
  # ...
  #
  # For every two elements read from memory, we perform 'n' additions.
  return y
\end{lstlisting}

\begin{lstlisting}[language=Python]
@functools.partial(jax.jit, static_argnames='steps')
def fast_exponentiation(x: jax.Array, steps: int):
  z = x
  for _ in range(1, steps):
    z = z * z
  return z
\end{lstlisting}

\begin{figure*}[h]
\includegraphics[width=0.9\linewidth]{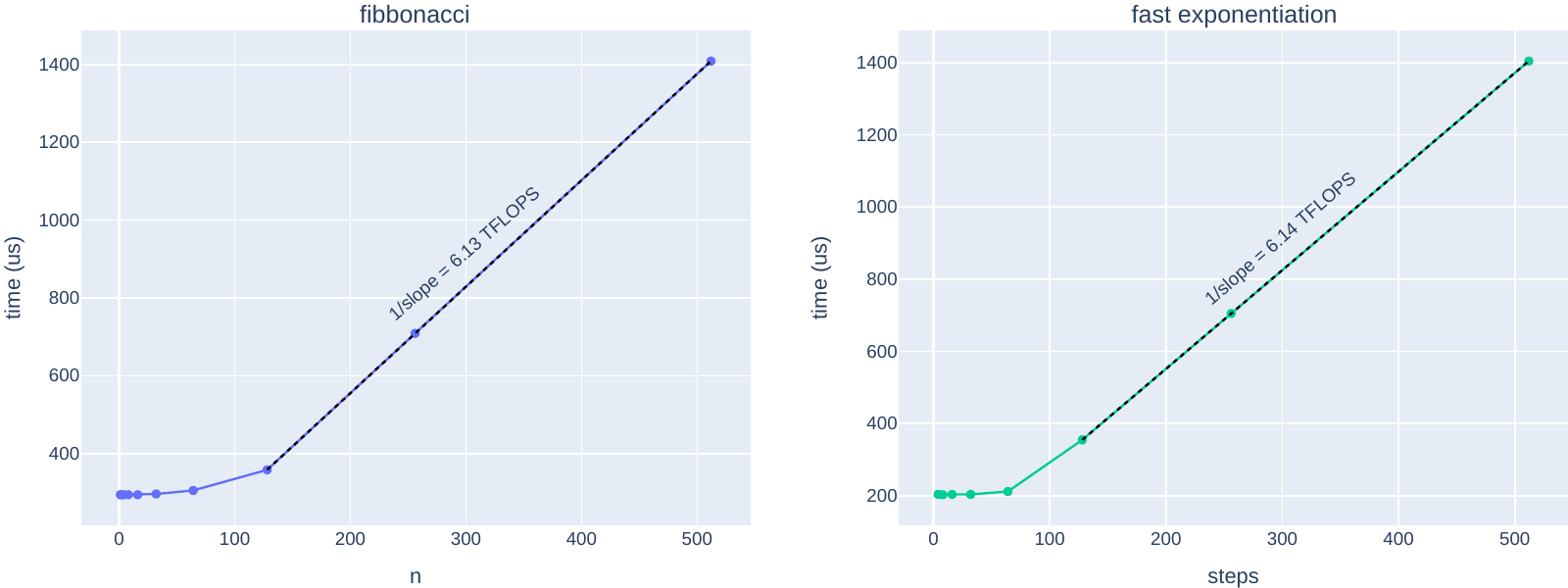}
\caption{\textbf{Estimating the throughput of the VPU on TPUv5e.} We expect the kernels to be memory-bound (constant line) initially and then be vector compute bound (linear scaling). We fit a line to the points in the linear region with the following model: $time = num\_ops \times \frac{1}{throughput} + overhead$. The inverse of the slope is an estimate of the peak throughput of the VPU.}
\end{figure*}

\FloatBarrier
\clearpage
\subsection{Visual illustration of the \citet{chern2022tpuknnknearestneighbor} algorithm.}
\begin{figure*}[h]
\includegraphics[width=\linewidth]{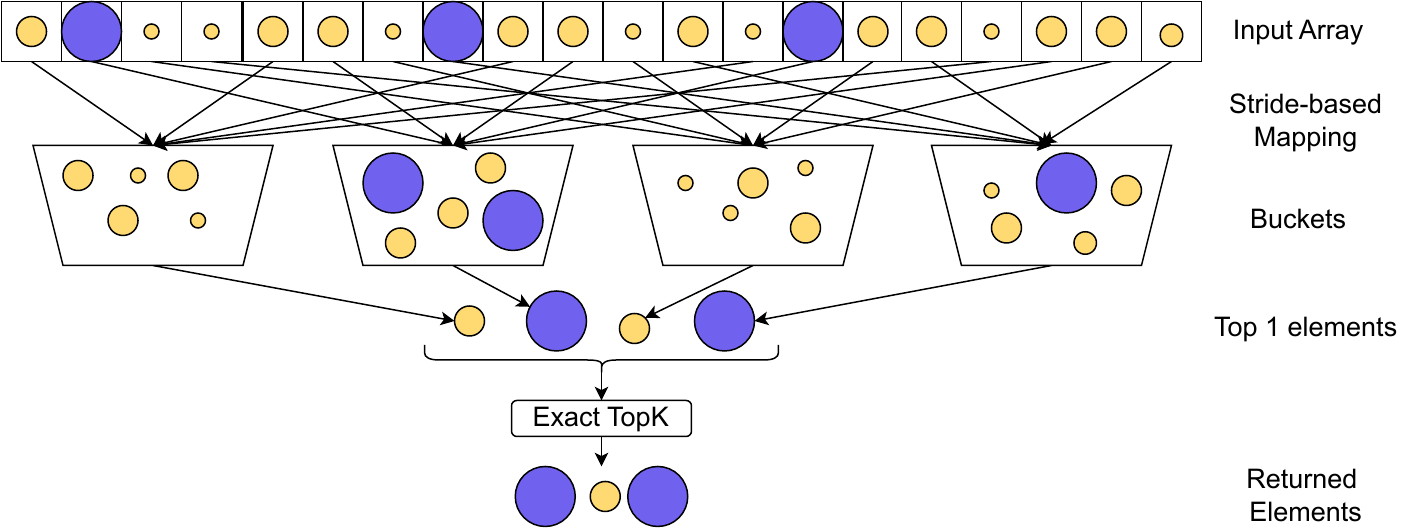}
\caption{\textbf{The two-stage approximate Top-$K$ algorithm by \citep{chern2022tpuknnknearestneighbor}.} This example demonstrates the process of finding the approximate top three elements from an array of twenty elements using the algorithm by~\citet{chern2022tpuknnknearestneighbor}. Ten buckets are required to guarantee an expected recall of 85\%, but we use only four for illustration purposes. The size of the balls indicates their value, and the top three balls have been colored blue for visual clarity. The first stage groups elements separated by a fixed stride of four into buckets and selects the top-1 element from each bucket. An exact Top-$K$ algorithm is applied on the selected elements to obtain the final result. In this example, two of the three actual top balls map to the same bucket, and one is dropped, resulting in an approximation error.}
\label{fig:top1algo}
\end{figure*}

\FloatBarrier
\clearpage
\subsection{Empirical verification of the quality of Monte Carlo estimates for expected recall.}
\label{ssec:mc_expr_verification}

\begin{figure*}[h]
\includegraphics[width=\linewidth]{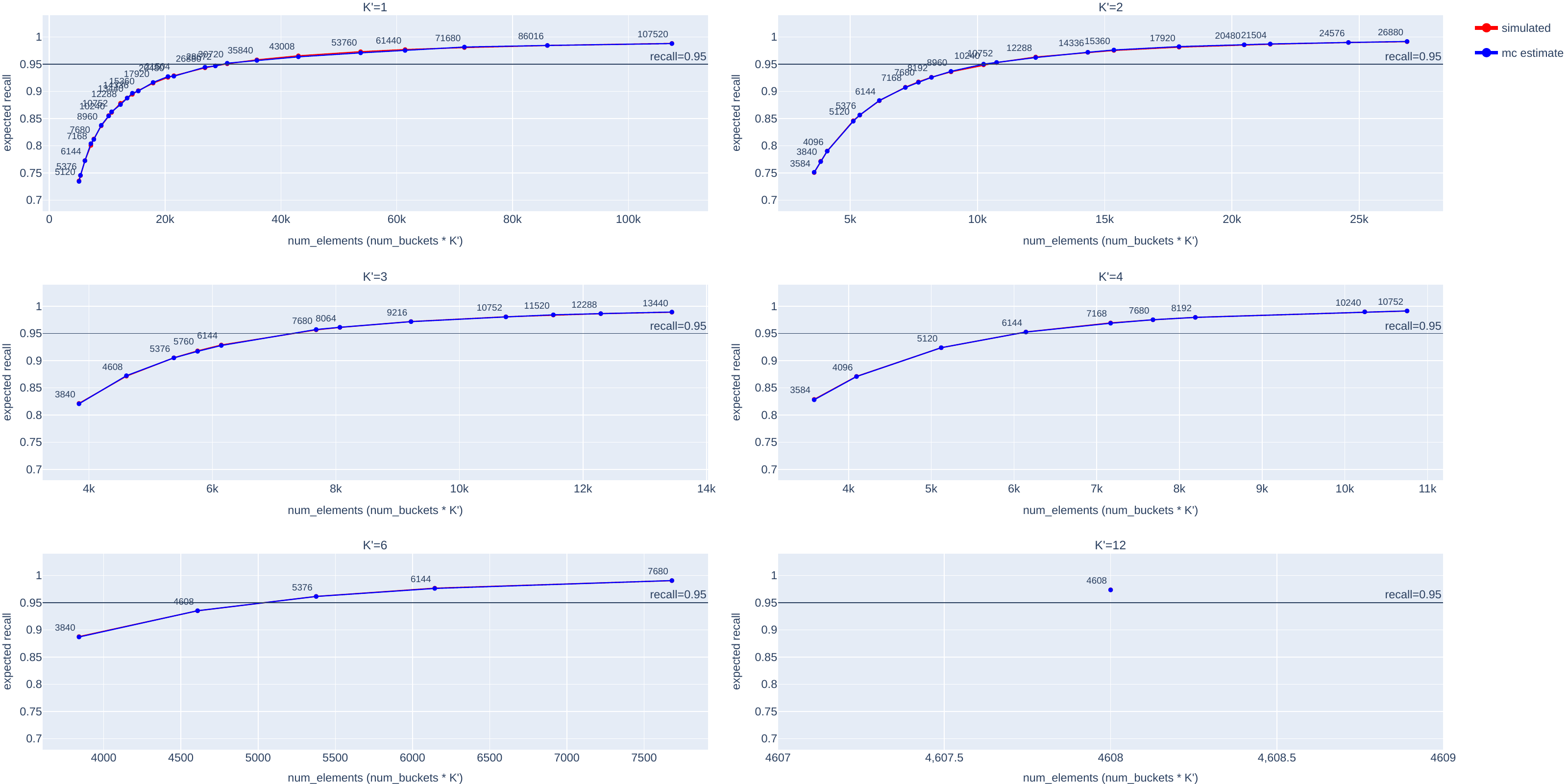}
\caption{\textbf{Verification of Monte Carlo estimates of expected recall against simulated runs of the algorithm for finding top-3360 ($\approx$ 0.8\%) elements from an array of size 430,080.} The simulated estimates were obtained from 1024 runs of the algorithm on randomly generated integers and the Monte Carlo estimates were obtained from 262144 samples of the expectation expression derived in Section \ref{sec:methods:analysis}.}
\end{figure*}

\begin{figure*}[h]
\includegraphics[width=\linewidth]{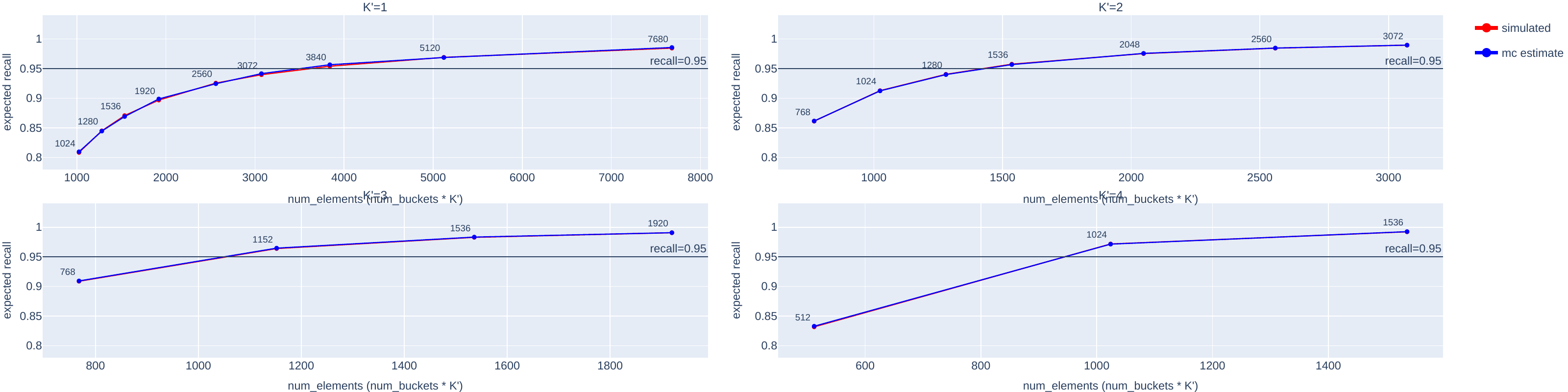}
\caption{\textbf{Verification of Monte Carlo estimates of expected recall against simulated runs of the algorithm for finding top-480 ($\approx$ 3\%) elements from an array of size 15,360.} The simulated estimates were obtained from 1024 runs of the algorithm on randomly generated integers and the Monte Carlo estimates were obtained from 262,144 samples of the expectation expression derived in Section \ref{sec:methods:analysis}.}
\end{figure*}

\FloatBarrier
\clearpage
\subsection{Proof of Theorem \ref{thm:exp_recall}.}
\label{asec:th1_proof}

\begin{proof}(Proof of Theorem~\ref{thm:exp_recall})
Consider an arbitrary subset $S \subseteq \{1,\cdots, N\}$ such that $|S| = K$. Let $S_j := S \cap G_j$ for $j=1,\cdots,B$. We now want to bound the number of elements in $S_j$ greater than $K'$ for some given $K'$:
\begin{align*}
m_j &:= 
\Ex{\max\left(0, \abs{S_j}-K'\right)} \\ &= \sum_{r=K'+1}^{\min(K,N/B)} (r-K') \frac{{K \choose r} {{N-K} \choose {\frac{N}{B}-r}}}{{N \choose {\frac{N}{B}}}},
\end{align*}
where each term in the summation refers to having $\abs{S_j}=r$; ${K \choose r}$ refers to choosing $r$ elements out of $S$; ${{N-K} \choose {\frac{N}{B}-r}}$ refers to the number of subsets, where $\frac{N}{B}-r$ elements in $G_j$ are chosen from outside of $S$; and ${N \choose {\frac{N}{B}}}$ refers to the total number of possible subsets that $G_j$ can take. Finally, the recall (i.e., the expected number of elements in $S$ eventually captured by the output of our algorithm) is given by:
\begin{align*}
    \Ex{\textrm{Recall}} = 1 - \frac{B \cdot m_j}{K}.
\end{align*}

We now show that the above expression is provably tighter than the expression obtained in~\citet{chern2022tpuknnknearestneighbor} for $K'=1$. Specifically, for $K'=1$, note that:
\begin{align}
    & \qquad \Ex{\abs{S_j}-K'} = -1 \cdot \Prob{\abs{S_j}=0} + m_j \\
    &\Rightarrow m_j = \Ex{\abs{S_j}} - 1 + \Prob{\abs{S_j}=0} \\
    &\Rightarrow m_j = \frac{K}{B} - 1 + \frac{{{N-K} \choose {\frac{N}{B}}}}{{N \choose \frac{N}{B}}} \\
    & \Rightarrow m_j \leq \frac{K}{B} - 1 + \left(1 - \frac{K}{N}\right)^{\frac{N}{B}} \\
    & \Rightarrow m_j \leq \frac{K}{B} - 1 + 1 - \frac{N}{B} \frac{K}{N} + {\frac{N}{B} \choose 2} \left(\frac{K}{N}\right)^2 \label{step:binomexp}\\
    & \Rightarrow m_j \leq \frac{K^2}{2 B}\left(\frac{1}{B} - \frac{1}{N}\right).
\end{align}
From the above, we see that the expected recall can be bounded as:
\begin{align*}
    \Ex{\textrm{Recall}} &\geq 1 - \frac{B}{K} \cdot \frac{K^2}{2 B}\left(\frac{1}{B} - \frac{1}{N}\right)\\
    &= 1 - \frac{K}{2}\left(\frac{1}{B} - \frac{1}{N}\right), 
\end{align*}
or equivalently, if we choose
\begin{align*}
    B = \frac{K}{2 \left(1- r + \frac{K}{2N}\right)},
\end{align*}
we will then have $\Ex{\textrm{Recall}} \geq r$. On the other hand,~\citet{chern2022tpuknnknearestneighbor} use $B = \frac{K}{1-r}$ to guarantee a recall of $r$, which is more than twice as large as required by our formula.

\[
    \underbrace{\frac{1}{2} \cdot\frac{K}{\left(1- r + \frac{K}{2N}\right)}}_{\text{our formula}} < \frac{1}{2}\cdot \left(\frac{K}{1- r}\right) < \underbrace{\frac{K}{1 - r}}_{\text{their formula}}
\]

In Appendix Section \ref{ssec:k1_bound_verification}, we verify the tightness of our bound and show that expanding up to the quartic term in step \ref{step:binomexp} provides a near-perfect approximation of the exact expression that is practically indistinguishable.
\end{proof}

\FloatBarrier
\clearpage
\subsection{Quality of the theoretical bounds on expected recall for $K'=1$ setting.}
\label{ssec:k1_bound_verification}

\begin{figure*}[h]
\includegraphics[width=\linewidth]{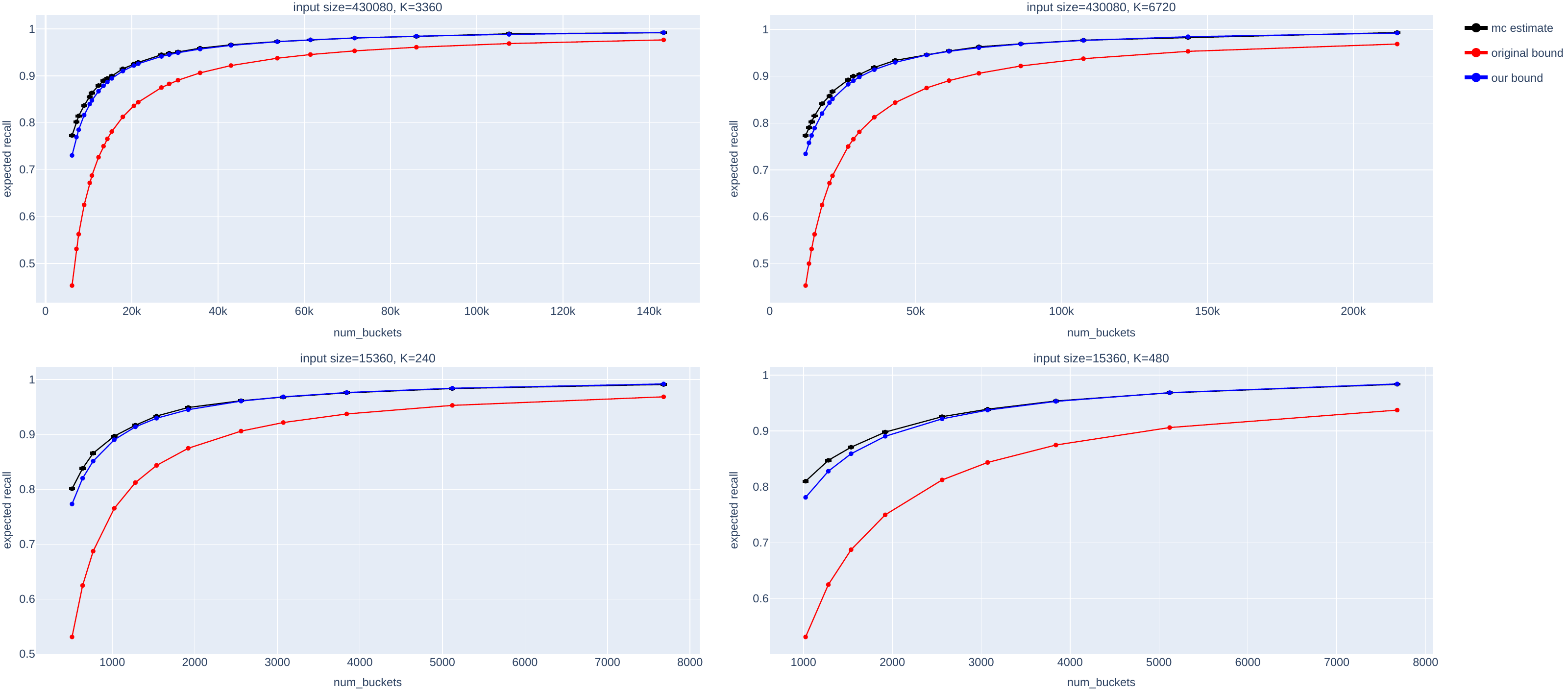}
\caption{\textbf{Tightness of our theoretical bound on expected recall for $K'=1$ setting compared to the original bound derived in \citet{chern2022tpuknnknearestneighbor}}. See Section \ref{sec:methods:analysis} for the derivation of our bound ($\Ex{\text{Recall}} \ge 1 - \frac{K}{2B}$) which is tighter than the original bound ($\Ex{\text{Recall}} \ge 1 - \frac{K}{B}$).}
\end{figure*}

\begin{figure*}[h]
\includegraphics[width=\linewidth]{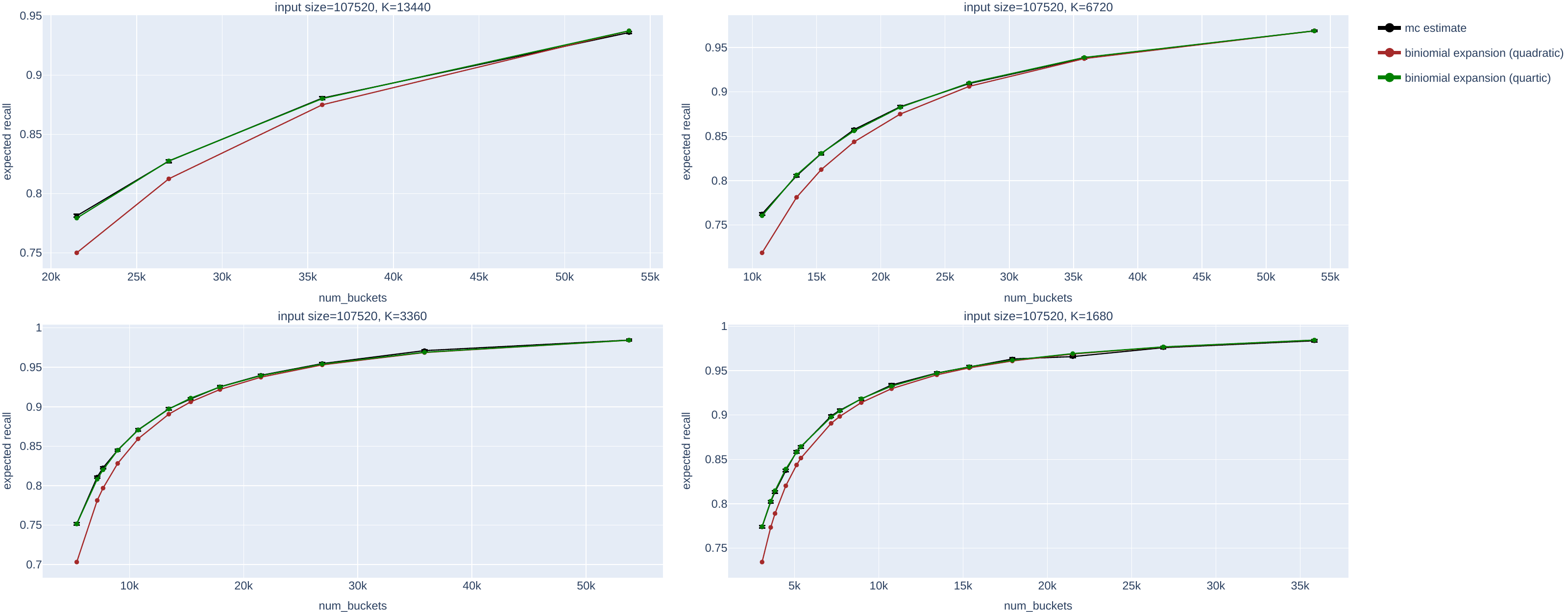}
\caption{\textbf{Expanding the binomial expression in Step \ref{step:binomexp} to quartic terms in Theorem \ref{thm:exp_recall} approximates the expected recall that is nearly exact.}}
\end{figure*}

\FloatBarrier
\clearpage
\subsection{Comparison to the work by \cite{key2024approximate}.}
\label{asec:concurrent_work}

We distinguish our theoretical analysis from the work of \citet{key2024approximate} primarily through the rigor and conceptual simplicity of our model. The recall expression in \citet{key2024approximate} (Appendix C) is approximate rather than exact. Specifically, because the algorithm distributes the $N$ elements into $B$ buckets, the distribution of the actual Top-$K$ elements in each bucket follows the hypergeometric distribution, not the binomial distribution. Therefore, their resulting expression is independent of $N$, which is only an asymptotic approximation. As shown in our Appendix Section \ref{ssec:k1_bound_verification}, the dependence on $N$ is non-negligible in low recall regimes; our exact derivation (Appendix Section \ref{ssec:mc_expr_verification}) captures this variance, leading to essentially tight bounds for the $K'=1$ case (Appendix Section \ref{asec:th1_proof}). Furthermore, our derivation is conceptually simpler and avoids state tracking or recursive expressions used in \cite{key2024approximate}.

\FloatBarrier
\clearpage
\subsection{Packages and utility functions for the code listings.}
\begin{lstlisting}[language=Python]
import jax
import jax.numpy as jnp
from jax.experimental import pallas as pl

import numpy as np

def get_all_factors(n):
  small_factors = [i for i in range(1, int(np.ceil(np.sqrt(n)))) if n % i == 0]
  pair_factors = [n // factor for factor in small_factors]
  return set(small_factors + pair_factors)
\end{lstlisting}

\subsection{Unfused implementation of our algorithm}
\begin{lstlisting}[language=Python, label=lst:pallas_implementation]
def generalized_partial_reduce(inputs, local_K, num_buckets, tunable_params={}, **kwargs):
  """ApproxTopK with generalized partial reduce for minor-axis reductions.

  Note: Input elements separated by a fixed stride form a bucket.

  Args:
    inputs: jax.ShapeDtypeStruct-compatible object of the input array of the
      shape [batch_size, reduction_dims].
    local_K: number of top elements to keep track of per bucket
    num_buckets: number of buckets
    tunable_params: These are hardware-specific (auto)tunable parameters. See
      the uses in the function body for more information.
    *kwargs: forwarded as is to `pallas_call`.

  Note: The default tunable params selection does not check if the choices are
    meaningful (e.g., VMEM OOMs). Please tune the choices if they don't work.

  Note: The procedure does not allow arbitrary input shapes for performance
    reasons and triggers assertions in case of incompatible shapes. Please pad
    the input shapes accordingly.

  Returns:
    A unary function implementing the algorithm that takes the input array and
    returns a tuple of TopK values and indices with the shape ([batch_size,
    num_elements], [batch_size, num_elements]) where num_elements is `local_K *
    num_buckets`.
  """

  # Pallas imposes constraints on block specifications. Refer to the docs.
  PALLAS_TPU_BLOCKSPEC_MAJOR_MULTIPLE = 8
  PALLAS_TPU_BLOCKSPEC_MINOR_MULTIPLE = 128

  input_shape = inputs.shape
  batch_size, reduction_dims = input_shape
  num_elements = num_buckets * local_K
  output_shape = (batch_size, num_elements)

  batch_tile_size = tunable_params.get('batch_tile_size', None)
  if batch_tile_size is None:
    factors = get_all_factors(batch_size)
    legal_factors = {
        f for f in factors
          if f % PALLAS_TPU_BLOCKSPEC_MAJOR_MULTIPLE == 0
          or f == batch_size
    }
    # Higher tile size provide more opportunities for instruction-level
    # parallelism.
    batch_tile_size = max({ f for f in legal_factors if f <= 2048 })
  assert(batch_size % batch_tile_size == 0)

  reduction_tile_size = tunable_params.get('reduction_tile_size', None)
  if reduction_tile_size is None:
    # In each subprogram, we would want to process sufficient number of inputs
    # to cover all the buckets. We would also want to have several passes over
    # the buckets in each subprogram so that the compiler can schedule the loop
    # iterations in a way that state loads/stores to the same buckets run
    # consecutively, and the state information is cached in registers or in the
    # nearest cache.
    factors = set(get_all_factors(reduction_dims))
    legal_factors = {
        f for f in factors
          if f % num_buckets == 0 and
             f % PALLAS_TPU_BLOCKSPEC_MINOR_MULTIPLE == 0
    }
    # We want to pick sufficiently large blocks so that the overheads are
    # amortized. However, we don't want the blocks to be too large to a point
    # that we have too few pipelined iterations and the head and tail latencies
    # make up a substantial portion of the runtime. These numbers vary from chip
    # to chip and need to be tuned.
    reduction_tile_size = max({
        d for d in legal_factors if d <= max(32_768, num_buckets)
    })
  assert(reduction_dims % reduction_tile_size == 0)

  # For simplicity, we restrict the tile sizes to be such that all the buckets
  # are processed equal number of times in each subprogram.
  assert(reduction_tile_size % num_buckets == 0)

  input_transform_indices_fn = lambda i, j: (i, j)
  input_tile_shape = (batch_tile_size, reduction_tile_size)
  iteration_bounds = [
      asz // tsz for asz, tsz in zip(input_shape, input_tile_shape)
  ]
  assert(len(input_tile_shape) == len(input_shape))

  # Pallas currently does not allow non-consecutive grid points to write to the
  # same slices of output. We therefore do not block the output along the
  # reduction axis.
  output_transform_indices_fn = lambda i, j: (i, 0)
  output_tile_shape = (batch_tile_size, num_elements)
  assert(len(output_tile_shape) == len(output_shape))

  # At the time of writing, the Mosaic compiler does not have a rule for
  # lowering comparisons between types that are not 32-bits wide. Hence, we
  # explicitly promote inputs to their wider 32-bit type.
  assert(inputs.dtype.itemsize <= 4)
  if jnp.issubdtype(inputs.dtype, jnp.floating):
    compute_type = jnp.float32
  elif jnp.issubdtype(inputs.dtype, jnp.signedinteger):
    compute_type = jnp.int32
  elif jnp.issubdtype(inputs.dtype, jnp.unsignedinteger):
    compute_type = jnp.uint32
  else:
    assert "Unknown data type"

  def _kernel(inputs_ref, values_ref, indices_ref):
    assert(values_ref.shape == indices_ref.shape)

    # b = batch axis, r = reduction axis
    tile_b, tile_r = pl.program_id(0), pl.program_id(1)

    # On TPUs, we are guaranteed a sequential grid execution and we use the
    # first run for each batch to initialize the outputs.
    @pl.when(tile_r == 0)
    def initialize_outputs():
      # We don't have to initialize the indices as non-strict comparators for
      # selection guarantee that the indices will be updated.
      values_ref[...] = jnp.full_like(values_ref, -jnp.inf)

    # The loop count may be large but we explicitly want to unroll to eliminate
    # a lot of state loads/stores. Rewriting the loop as two nested loops where
    # the unrolled inner loop explicitly reuses the state load/stores and the
    # non-unrolled outer loop runs over different sets of buckets may lead to
    # faster compilation.
    num_iterations_over_outputs = reduction_tile_size // num_buckets
    for iter_idx in range(num_iterations_over_outputs):
      # Note that inputs are already tiled by pallas and we use local offsets.
      inputs = inputs_ref[
          :, pl.ds(start=iter_idx * num_buckets, size=num_buckets)
      ]
      inputs = inputs.astype(compute_type)

      iota = jax.lax.broadcasted_iota(indices_ref.dtype, inputs.shape, 1)
      iota += tile_r * reduction_tile_size + iter_idx * num_buckets
      assert(inputs.shape == iota.shape)

      # Load state information for the current chunk.
      values_by_k, indices_by_k = [], []
      for k in range(local_K):
        values = values_ref[
            :,
            pl.ds(
                start=k * num_buckets,
                size=num_buckets
            )
        ].astype(compute_type)
        indices = indices_ref[
            :,
            pl.ds(
                start=k * num_buckets,
                size=num_buckets
            )
        ]
        assert(values.shape == indices.shape)
        values_by_k.append(values)
        indices_by_k.append(indices)

      # Compute the new state information for the current chunk.
      pred = inputs >= values_by_k[-1]
      values_by_k[-1] = jax.lax.select(pred, inputs, values_by_k[-1])
      indices_by_k[-1] = jax.lax.select(pred, iota, indices_by_k[-1])
      for k in reversed(range(1, local_K)):
        # Note that the commented line and uncommented line are algorithmically
        # equivalent, but the uncommented version has one less loop-carried
        # dependency.
        # pred = values_by_k[k] > values_by_k[k - 1]
        pred = inputs > values_by_k[k - 1]

        values_to_shift = values_by_k[k]
        values_by_k[k] = \
          jax.lax.select(pred, values_by_k[k-1], values_to_shift)
        values_by_k[k-1] = \
          jax.lax.select(pred, values_to_shift, values_by_k[k-1])

        indices_to_shift = indices_by_k[k]
        indices_by_k[k] = \
          jax.lax.select(pred, indices_by_k[k-1], indices_to_shift)
        indices_by_k[k-1] = \
          jax.lax.select(pred, indices_to_shift, indices_by_k[k-1])

      # Write the new state information for the current chunk.
      for k in range(local_K):
        values_ref[
            :,
            pl.ds(
                start=k * num_buckets,
                size=num_buckets
            )
        ] = values_by_k[k].astype(values_ref.dtype)
        indices_ref[
            :,
            pl.ds(
                start=k * num_buckets,
                size=num_buckets
            )
        ] = indices_by_k[k]

  def wrapper(inputs):
    pr_values, pr_indices = pl.pallas_call(
        _kernel,
        in_specs=[
            pl.BlockSpec(input_tile_shape, input_transform_indices_fn),
        ],
        out_shape=[
            jax.ShapeDtypeStruct(output_shape, inputs.dtype),
            jax.ShapeDtypeStruct(output_shape, jnp.int32),
        ],
        out_specs=[
            pl.BlockSpec(output_tile_shape, output_transform_indices_fn),
            pl.BlockSpec(output_tile_shape, output_transform_indices_fn)
        ],
        grid=iteration_bounds,
        compiler_params=pltpu.TPUCompilerParams(
          dimension_semantics=("parallel", "arbitrary")
        ),
        **kwargs
    )(inputs)
    return pr_values, pr_indices
  return wrapper
\end{lstlisting}

\begin{lstlisting}[language=Python]
def make_generalized_approx_topk(operand, num_buckets, local_K, global_K, **kwargs):
    partial_reduce_fn = \
      generalized_partial_reduce(operand, local_K, num_buckets, **kwargs)

    def wrapper(operand):
      bucket_values, bucket_indices = partial_reduce_fn(operand)
      values, indices = \
        jax.lax.sort_key_val(bucket_values, bucket_indices, is_stable=False)
      values = jnp.flip(values[..., -global_K:], axis=-1)
      indices = jnp.flip(indices[..., -global_K:], axis=-1)
      return values, indices
    return wrapper
\end{lstlisting}

\subsection{Matmul-fused implementation of our algorithm.}
\begin{lstlisting}[language=Python, label=lst:fused_pallas_implementation]
def matmul_fused_generalized_partial_reduce(
    lhs, rhs,
    local_K, num_buckets,
    tunable_params={}, *, **kwargs
):
  """Fused ApproxTopK with generalized partial reduce for minor-axis reductions.

  Note: Input elements separated by a fixed stride form a bucket.

  Args:
    lhs: jax.ShapeDtypeStruct-compatible object of LHS array with shape
      [batch_size, contracting_dims].
    rhs: jax.ShapeDtypeStruct-compatible object of RHS array with shape
      [contracting_dims, reduction_dims].
    local_K: number of top elements to keep track of per bucket.
    num_buckets: number of buckets.
    tunable_params: These are hardware-specific (auto)tunable parameters. See
      the uses in the function body for more information.
    **kwargs: forwarded as is to `pallas_call`.

  Note: The default tunable params selection does not check if the choices are
    meaningful (e.g., VMEM OOMs). Please tune the choices if they don't work.

  Note: The procedure does not allow arbitrary input shapes for performance
    reasons and triggers assertions in case of incompatible shapes. Please pad
    the input shapes accordingly.

  Returns:
    A binary function implementing the algorithm that takes the arguments to
    matmul and returns a tuple of TopK values and indices of lhs @ rhs with
    shapes of ([batch_size, num_elements], [batch_size, num_elements]).
    The `num_elements` is calculated as `num_buckets * local_K`.
  """

  # Pallas imposes constraints on block specifications. Refer to the docs.
  PALLAS_TPU_BLOCKSPEC_MAJOR_MULTIPLE = 8
  PALLAS_TPU_BLOCKSPEC_MINOR_MULTIPLE = 128

  # This implementation maps buckets to the minormost axis.
  assert(num_buckets % PALLAS_TPU_BLOCKSPEC_MINOR_MULTIPLE == 0)

  batch_size, contracting_dims = lhs.shape
  contracting_dims_rhs, reduction_dims = rhs.shape
  assert(contracting_dims == contracting_dims_rhs)
  assert(reduction_dims % num_buckets == 0)
  assert(num_buckets < reduction_dims)
  assert(lhs.dtype == rhs.dtype)

  num_elements = num_buckets * local_K
  output_shape = (batch_size, num_elements)

  # We will block the matrices for software pipelining as follows:
  # lhs: [batch_tile_size, contracting_tile_size]
  # rhs: [contracting_tile_size, reduction_tile_size]
  # result-scratch: [batch_tile_size, reduction_tile_size]
  #
  # Note that partial reduce computation can start only after the loop over the
  # contracting axis ends, as it requires fully accumulated sums to begin.
  # The VPU may idle waiting for the result tile in all but the last iteration
  # of the loop over the contracting axis.
  #
  # We can alleviate the problem by pipelining the computation into matmul and
  # TopK stages. We let the VPU processes the previous result tile while the new
  # result tile is being accumulated. We do not implement the idea here and
  # choose to use large tiling along contracting axis to minimize wasted cycles.

  # The tiles will be further subtiled automatically by the compiler to meet the
  # shape of the hardware matmul units. Since the subtiling loops will be fully
  # unrolled, the compiler would ideally generate code to run TopK on the
  # previous subtile while a new subtile is being accumulated.

  batch_tile_size = tunable_params.get('batch_tile_size', None)
  if batch_tile_size is None:
    # We want this to be as large as possible. This parameter controls the
    # arithmetic intensity of the blocked matmul operation. Therefore, we must
    # have a batch tile size that is high enough to ensure that each blocked
    # matmul operation is MXU-bound.
    factors = get_all_factors(batch_size)
    legal_factors = {
        f for f in factors
          if f % PALLAS_TPU_BLOCKSPEC_MAJOR_MULTIPLE == 0
          or f == batch_size
    }
    # We heuristically pick the largest legal tile size up to 2048. A larger
    # tile size may be more performant but carries the risk of exhausting VMEM.
    batch_tile_size = max({ f for f in legal_factors if f <= 2048 })
  assert(batch_size % batch_tile_size == 0)

  contracting_tile_size = tunable_params.get('contracting_tile_size', None)
  if contracting_tile_size is None:
    # This tile size does not affect the arithmetic intensity of the matrix
    # multiplication. However, as mentioned earlier, we cannot start the TopK
    # computation without the fully accumulated result tile. To minimize VPU
    # idle time, we would like to have as large a tile size as possible for the
    # contracting axis so that there are as few iterations as possible where the
    # final result tile is only partially accumulated.
    factors = get_all_factors(contracting_dims)

    # This axis would be the minor axis for LHS and the major axis for RHS. It
    # must meet the multiple requirements for the LHS and RHS respectively or
    # must be equal to the axis size.
    legal_factors = {
        f for f in factors
          if (f % PALLAS_TPU_BLOCKSPEC_MAJOR_MULTIPLE == 0 and
              f % PALLAS_TPU_BLOCKSPEC_MINOR_MULTIPLE == 0)
          or f == contracting_dims
    }

    # We heuristically pick the largest size up to 2048. A larger tile size
    # would minimize VPU idling for this implementation but increases the risk
    # of VMEM OOMs.
    contracting_tile_size = max({ f for f in legal_factors if f <= 2048 })
  assert(contracting_dims % contracting_tile_size == 0)

  reduction_tile_size = tunable_params.get('reduction_tile_size', None)
  if reduction_tile_size is None:
    # There are two possibilities for picking reduction tile size:
    # 1. tile size > number of buckets
    # 2. tile size <= number of buckets
    #
    # Our implementation handles both cases. However, the first possibility is
    # preferred for performance reasons.

    # We need to ensure that we have sufficient VMEM to accommodate the large
    # tiles for matrix multiplication so that it remains MXU-bound. Let's cap
    # the tile size to 4096 to reduce the risk of exhausting VMEM.
    if num_buckets > 4096:
      assert(reduction_dims % num_buckets == 0)
      factors = set(get_all_factors(num_buckets))
      legal_factors = {
          f for f in factors
            if f % PALLAS_TPU_BLOCKSPEC_MINOR_MULTIPLE == 0
      }
      reduction_tile_size = max({ d for d in legal_factors if d <= 4096 })
    else:
      # We want to pick sufficiently large tiles so that the load/store overhead
      # of the TopK lists is amortized. However, we don't want the blocks to be
      # too large to a point that we have too few pipelining iterations and the
      # head and tail latencies make up a substantial portion of the runtime.
      assert(reduction_dims % num_buckets == 0)
      factors = set(get_all_factors(reduction_dims))
      legal_factors = {
          f for f in factors
          if f % num_buckets == 0
          and f % PALLAS_TPU_BLOCKSPEC_MINOR_MULTIPLE == 0
      }
      reduction_tile_size = max({ f for f in legal_factors if f <= 4096 })

  assert(reduction_dims % reduction_tile_size == 0)
  if reduction_tile_size > num_buckets:
    # For simplifying the implementation, we restrict the tile size to be
    # multiples of number of buckets.
    assert(reduction_tile_size % num_buckets == 0)
  else:
    # For simplifying the implementation, we restrict the tile size to be
    # factors of number of buckets.
    assert(num_buckets % reduction_tile_size == 0)

  lhs_transform_indices_fn = lambda i, j, k: (i, k)
  lhs_tile_shape = (batch_tile_size, contracting_tile_size)
  assert(len(lhs_tile_shape) == len(lhs.shape))

  rhs_transform_indices_fn = lambda i, j, k: (k, j)
  rhs_tile_shape = (contracting_tile_size, reduction_tile_size)
  assert(len(rhs_tile_shape) == len(rhs.shape))

  result_tile_shape = (batch_tile_size, reduction_tile_size)

  iteration_bounds = [asz // tsz for asz, tsz in zip(
      [batch_size, reduction_dims, contracting_dims],
      [batch_tile_size, reduction_tile_size, contracting_tile_size]
  )]
  contraction_steps = iteration_bounds[2]

  # Pallas currently does not allow non-consecutive grid points to write to the
  # same slices of output. We therefore do not block the output along the
  # reduction axis.
  output_transform_indices_fn = lambda i, j, k: (i, 0)
  output_tile_shape = (batch_tile_size, num_elements)
  assert(len(output_tile_shape) == len(output_shape))

  # At the time of writing, the Mosaic compiler does not have a rule for
  # lowering comparisons between types that are not 32-bits wide. Hence, we
  # explicitly promote inputs to their wider 32-bit type.
  assert(lhs.dtype.itemsize <= 4)
  if jnp.issubdtype(lhs.dtype, jnp.floating):
    compute_type = jnp.float32
  elif jnp.issubdtype(lhs.dtype, jnp.signedinteger):
    compute_type = jnp.int32
  elif jnp.issubdtype(lhs.dtype, jnp.unsignedinteger):
    compute_type = jnp.uint32
  else:
    assert "Unknown data type"

  def _kernel(lhs_ref, rhs_ref, values_ref, indices_ref, acc_ref):
    # b = batch axis, r = reduction axis, c = contracting axis
    tile_b, tile_r, tile_c = \
      pl.program_id(0), pl.program_id(1), pl.program_id(2)

    if contraction_steps > 1:
      @pl.when(tile_c == 0)
      def reset_accumulators():
        acc_ref[...] = jnp.zeros_like(acc_ref)

      # For each output tile, we reset the accumulators to zero.
      @pl.when(tile_c < contraction_steps - 1)
      def matmul_only_step():
        acc_ref[...] += jnp.matmul(
            lhs_ref[...], rhs_ref[...], preferred_element_type=jnp.float32
        )

    # When we've accumulated all the partial products, we update the top-K'
    # lists with the new elements.
    @pl.when(tile_c == contraction_steps - 1)
    def update_topk_state():
      assert(values_ref.shape == indices_ref.shape)

      @pl.when(tile_r == 0)
      def initialize_outputs():
        # We don't have to initialize the indices as non-strict comparators for
        # selection guarantee that the indices will be updated.
        values_ref[...] = jnp.full_like(values_ref, -jnp.inf)

      if contraction_steps == 1:
        acc_ref[...] = jnp.zeros_like(acc_ref)

      acc_ref[...] += jnp.matmul(
          lhs_ref[...], rhs_ref[...], preferred_element_type=jnp.float32
      )

      def update_state(inputs, iota, state_offset, state_size):
        """Update the top-K' lists with the new inputs."""

        assert(inputs.shape == iota.shape)
        assert(inputs.shape[-1] == state_size)
        assert(state_size <= num_buckets)

        # Load state information for the current chunk.
        values_by_k, indices_by_k = [], []
        for k in range(local_K):
          values = values_ref[
              :,
              pl.ds(
                  start=pl.multiple_of(k * num_buckets + state_offset, 128),
                  size=state_size
              )
          ].astype(compute_type)
          indices = indices_ref[
              :,
              pl.ds(
                  start=k * num_buckets + state_offset,
                  size=state_size
              )
          ]
          assert(values.shape == indices.shape)
          values_by_k.append(values)
          indices_by_k.append(indices)

        # Compute the new state information for the current chunk.
        pred = inputs >= values_by_k[-1]
        values_by_k[-1] = jax.lax.select(pred, inputs, values_by_k[-1])
        indices_by_k[-1] = jax.lax.select(pred, iota, indices_by_k[-1])
        for k in reversed(range(1, local_K)):
          # The commented line and uncommented line are algorithmically
          # equivalent, but the uncommented version has one less loop-carried
          # dependency.
          # pred = values_by_k[k] > values_by_k[k - 1]
          pred = inputs > values_by_k[k - 1]

          values_to_shift = values_by_k[k]
          values_by_k[k] = \
            jax.lax.select(pred, values_by_k[k-1], values_to_shift)
          values_by_k[k-1] = \
            jax.lax.select(pred, values_to_shift, values_by_k[k-1])

          indices_to_shift = indices_by_k[k]
          indices_by_k[k] = \
            jax.lax.select(pred, indices_by_k[k-1], indices_to_shift)
          indices_by_k[k-1] = \
            jax.lax.select(pred, indices_to_shift, indices_by_k[k-1])

        # Write the new state information for the current chunk.
        for k in range(local_K):
          values_ref[
              :,
              pl.ds(
                  start=k * num_buckets + state_offset,
                  size=state_size
              )
          ] = values_by_k[k].astype(values_ref.dtype)
          indices_ref[
              :,
              pl.ds(
                  start=k * num_buckets + state_offset,
                  size=state_size
              )
          ] = indices_by_k[k]

      if reduction_tile_size > num_buckets:
        assert(reduction_tile_size % num_buckets == 0)
        num_iterations_over_outputs = reduction_tile_size // num_buckets

        # The loop count may be large but we explicitly want to unroll to
        # eliminate a lot of state loads/stores.
        for iter_idx in range(num_iterations_over_outputs):
          # The inputs are already tiled by pallas and we use local offsets.
          inputs = acc_ref[
              :, pl.ds(start=iter_idx * num_buckets, size=num_buckets)
          ].astype(compute_type)

          iota = jax.lax.broadcasted_iota(indices_ref.dtype, inputs.shape, 1)
          iota += tile_r * reduction_tile_size + iter_idx * num_buckets
          assert(inputs.shape == iota.shape)

          update_state(inputs, iota, 0, num_buckets)
      else:
        assert(num_buckets % reduction_tile_size == 0)
        inputs = acc_ref[...].astype(compute_type)

        iota = jax.lax.broadcasted_iota(indices_ref.dtype, inputs.shape, 1)
        iota += tile_r * reduction_tile_size
        assert(inputs.shape == iota.shape)

        num_tiles_over_outputs = num_buckets // reduction_tile_size
        state_offset = (tile_r % num_tiles_over_outputs) * reduction_tile_size

        update_state(inputs, iota, state_offset, reduction_tile_size)

  def wrapper(lhs, rhs):
    pr_values, pr_indices = pl.pallas_call(
      _kernel,
      grid_spec=pltpu.PrefetchScalarGridSpec(
        num_scalar_prefetch=0,
        in_specs=[
          pl.BlockSpec(lhs_tile_shape, lhs_transform_indices_fn),
          pl.BlockSpec(rhs_tile_shape, rhs_transform_indices_fn),
        ],
        out_specs=[
          pl.BlockSpec(output_tile_shape, output_transform_indices_fn),
          pl.BlockSpec(output_tile_shape, output_transform_indices_fn)
        ],
        scratch_shapes=[pltpu.VMEM(result_tile_shape, jnp.float32)],
        grid=iteration_bounds,
      ),
      out_shape=[
        jax.ShapeDtypeStruct(output_shape, compute_type),
        jax.ShapeDtypeStruct(output_shape, jnp.int32),
      ],
      compiler_params=pltpu.TPUCompilerParams(
        dimension_semantics=("parallel", "arbitrary", "arbitrary")
      ),
      **kwargs
    )(lhs, rhs)
    return pr_values, pr_indices
  return wrapper
\end{lstlisting}

\begin{lstlisting}[language=Python]
def make_matmul_fused_generalized_approx_topk(
    lhs, rhs, num_buckets, local_K, global_K, **kwargs
):
    partial_reduce_fn = \
      matmul_fused_generalized_partial_reduce(lhs, rhs, local_K, num_buckets, **kwargs)

    def wrapper(lhs, rhs):
      bucket_values, bucket_indices = partial_reduce_fn(lhs, rhs)
      values, indices = \
        jax.lax.sort_key_val(bucket_values, bucket_indices, is_stable=False)
      values = jnp.flip(values[..., -global_K:], axis=-1)
      indices = jnp.flip(indices[..., -global_K:], axis=-1)
      return values, indices
    return wrapper

def matmul_fused_generalized_approx_topk(lhs, rhs, *args, **kwargs):
  return make_matmul_fused_generalized_approx_topk(
      lhs, rhs, *args, **kwargs)(lhs, rhs)
\end{lstlisting}

\subsection{Algorithm Parameter Selection}
\label{lst:parameter_selection}

\subsubsection{Monte Carlo Estimation of Expected Recall}
\begin{lstlisting}[language=Python]
def expected_recall_mc(N, B, K_global, K_local, num_trials):
  assert(N % B == 0)
  bucket_size = N // B
  X_samples = np.random.hypergeometric(
      K_global,
      N - K_global,
      bucket_size,
      size=num_trials
  )
  num_collisions = B * np.maximum(X_samples - K_local, 0)
  recall = 1 - num_collisions / K_global
  expected_recall = np.mean(recall)
  std_error = np.std(recall, ddof=1) / np.sqrt(num_trials)
  return expected_recall, std_error
\end{lstlisting}

\subsubsection{Parameter Sweep}
\label{ssec:param_sweep}
\begin{lstlisting}[language=Python]
def select_parameters(
    input_size, K,
    recall_target,
    allowed_local_K=[1, 2, 3, 4]
  ):
  """Finds a good set of algorithm parameters for the given configuration.

  Args:
    input_size: size of the array
    K: number of top entries
    recall_target: minimum "expected" recall required
    allowed_local_K: list of local K to consider in the search space
  """

  divisors = get_all_factors(input_size)
  allowed_num_buckets = [ d for d in divisors if d % 128 == 0 ]

  # For a fixed K, the expected recall decreases as the number of buckets
  # decreases. Therefore, by sweeping through `num_buckets` in descending
  # order, we can terminate the search early when we miss the recall target.
  allowed_num_buckets = sorted(allowed_num_buckets, reverse=True)

  # The best configuration selection logic only checks for the total number of
  # elements using a strict comparison. We want to try local K in ascending
  # order so that in the case of a tie, we pick the configuration with a smaller
  # local K. 
  allowed_local_K = sorted(allowed_local_K)

  best_config = None
  best_num_elements = np.inf
  for local_K in allowed_local_K:
    for num_buckets in allowed_num_buckets:
      if num_buckets * local_K < K:
        break

      if recall_target >= 0.995:
        warnings.warn(
            f"recall_target of {recall_target} too high"
            " for reliable selection of algorithm.",
            RuntimeWarning
        )

      num_trials = 4096
      recall, recall_err = \
        expected_recall_mc(input_size, num_buckets, K, local_K, num_trials)
      while recall_err * 3 > 0.005:
        num_trials *= 2
        recall, recall_err = \
          expected_recall_mc(input_size, num_buckets, K, local_K, num_trials)

      if recall < recall_target:
        break

      num_elements = num_buckets * local_K
      if num_elements < best_num_elements:
        best_config = (local_K, num_buckets)
        best_num_elements = num_elements
  assert(best_config is not None)
  return best_config
\end{lstlisting}

\subsubsection{Computational cost of parameter selection routine.}
The parameter sweep evaluates configurations in descending order of bucket count for each value of $K'$, terminating early once the recall target is no longer met. This works since recall decreases monotonically with fewer buckets. For each configuration evaluated, we use an adaptive sampling procedure that draws hypergeometric samples until the recall estimate is within $\pm 0.005$ at 3-sigma confidence. To give a sense of the cost, we evaluated eight representative configurations spanning array sizes from 16k to 917k and K from 128 to 3360 at 95\% recall target; the parameter sweep evaluated between a few dozen and up to 115 configurations per (N, K, recall\_target) combination, drawing between 930k and 5M hypergeometric samples in total, and completed in under a second on an AMD Ryzen 7 3700X desktop CPU. This cost is negligible relative to the typical compilation times for LLMs, which range from a few minutes to several hours. In practice, the selected parameters can be cached and reused across all calls with the same configuration, which is common in LLMs where repeated identical blocks of layers dominate the architecture. The implementation has not been optimized; straightforward improvements such as binary search over bucket counts or sharing random samples across configurations could further reduce the cost.

\subsubsection{Discussion and Generalization}

The parameter sweep is currently specific to TPUv5e. The model in Section~\ref{sec:runtime_perf_analysis} and Table~\ref{tbl:hw_throughput} determines the maximum $K'$ for which the first stage remains memory-bound and therefore costs approximately the same regardless of $K'$ (confirmed empirically in Table~\ref{tbl:results_top1024_262144}, where Stage 1 latency is nearly flat from $K'=1$ to $K'=6$). Within this regime, minimizing the number of output elements ($B \times K'$) is equivalent to minimizing total latency, so the sweep correctly optimizes the right objective without explicitly consulting Table~\ref{tbl:hw_throughput} at search time. On TPUv5e, the ridge point analysis gives $K'=6$ as the crossover between memory-bound and compute-bound first-stage behavior; we conservatively restrict to $K' \leq 4$ to leave headroom for other operations like casting. The constraint that the number of buckets be a multiple of 128 and a divisor of the input size arises from TPUv5e-specific alignment requirements in our implementation.

The underlying methodology nonetheless generalizes. For any accelerator for which the quantities in Table~\ref{tbl:hw_throughput} can be measured or obtained from the datasheets, one can: (i) compute the ridge point to determine the $K'$ ceiling below which the first stage remains memory-bound; (ii) identify the implementation constraints; and (iii) run the same sweep to find the configuration minimizing second-stage input size. We can in principle extend this existing sweep routine to other TPUs directly using corresponding numbers from Table~\ref{tbl:hw_throughput}.

The benefits of hardware-aware parameter selection are most significant in fused settings. When operations such as ReLU, sigmoid, or elementwise products (e.g., in GLU-based MLP blocks) are fused alongside the Top-$K$ first stage, they compete for VPU instructions and reduce the effective vector compute budget available for Top-$K$, thereby lowering the effective $K'$ ceiling. Conversely, when fusing with matrix multiplications with large contracting dimensions, substantially more VPU compute becomes available than the baseline estimate from Section~\ref{sec:runtime_perf_analysis} suggests (as discussed in Section~\ref{sec:original_algorithm}), potentially enabling $K'$ well beyond 6. A complete fusion-aware parameter selection procedure would account for these competing demands when determining the $K'$ ceiling, and could in principle be integrated into a compiler cost model to automate fusion decisions. This framework applies broadly, not just to matmul fusions, but to any kernel with which Top-$K$ might be fused, including memory-bound kernels.

\clearpage
\FloatBarrier
\subsection{Expected recall rapidly improves with increasing $K'$.}
\label{lst:curve_separation}

\begin{figure}[!htb]
\includegraphics[width=\linewidth]{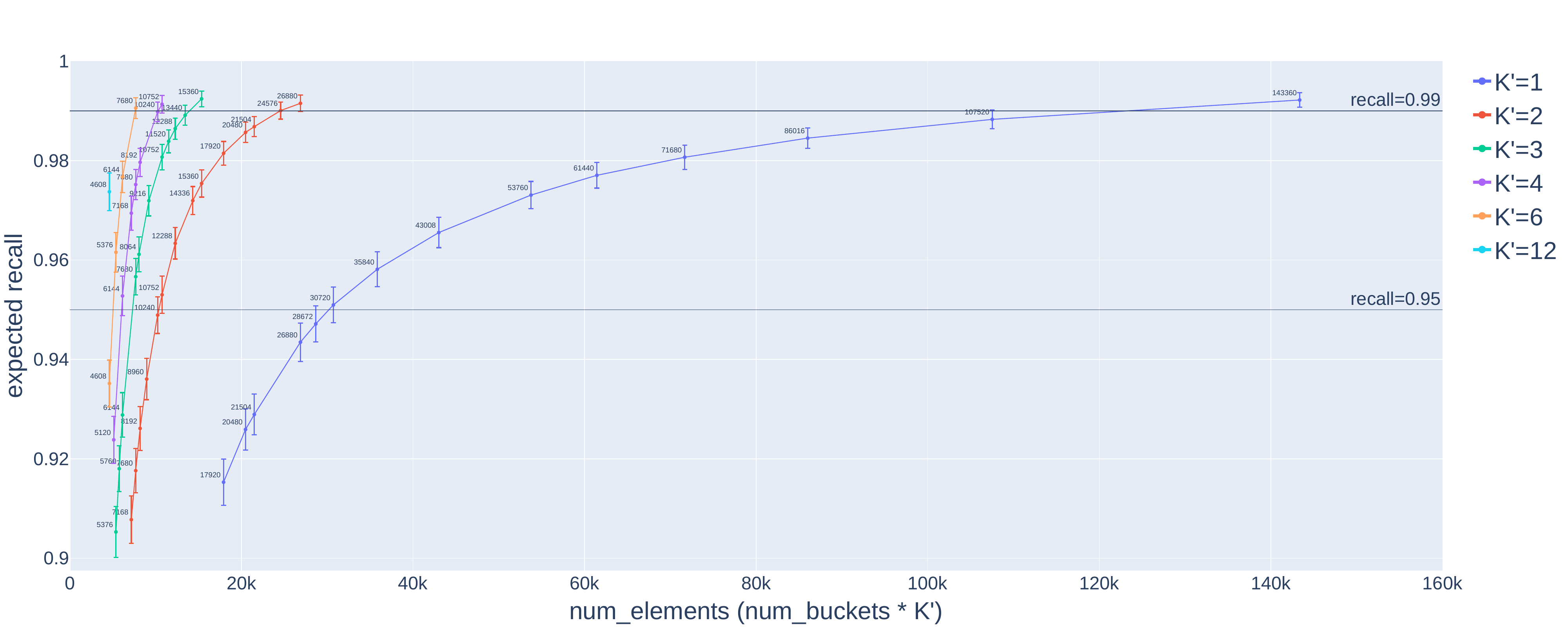}
\caption{\textbf{Recall vs number of elements for finding top-3360 ($\approx$ 0.8\%) elements from an array of size 430,080.} The data was obtained from simulated runs of the algorithm on randomly generated integers. The markers represent the sample mean and the error bars represent the sample standard deviation from 1024 trials. Each curve corresponding to a $K'$ depicts the Pareto frontier for that $K'$. The ideal point is $(K, 1.0)$. Beyond a certain $K'$, the first stage will become sufficiently expensive that the additional cost of the first stage exceeds the gains in the second stage. However, for small values of $K'$, where the additional cost of the first stage is negligible, we note that the Pareto frontier improves as $K'$ increases.}
\label{fig:accuracy_plot}
\end{figure}

\FloatBarrier
\subsection{Arithmetic intensity of fused matmul operation.}
\label{sec:aimatmul_derive}

\begin{align*}
    \text{arithmetic intensity} &= \frac{2BDN}{E\left(BD + DN + BN\right)}\\
    &\approx \frac{2BDN}{E\left(DN + BN\right)}\\
    &= \frac{2BD}{E\left(B + D\right)}\\
    &\le \frac{2}{E}\min(B, D).
\end{align*}

\FloatBarrier
\subsection{Training for sparse activations in MLP blocks.}
\label{sec:speedups_sparse_mlp_training}

We benchmark a non-gated MLP variant of Gemma 2 9B using SquaredReLU activations to induce sparsity, following prior work on activation sparsity \citep{l2024hirehighrecallapproximate, mirzadeh2023relustrikesbackexploiting, zhang2024relu2winsdiscoveringefficient}. To maintain rough parameter parity with the gated baseline, we set the intermediate dimension in the MLP blocks to 24,576, while keeping all other architectural hyperparameters unchanged. We use a sequence length of 1024 and a per-rank batch size of 8. To enforce sparsity, we apply a Top-K algorithm that selects approximately the top 2\% of FFN activations (K = 512 out of 24,576, i.e., 2.08\%), targeting 95\% recall.

We profile isolated residual blocks (including both forward and backward passes). In the dense baseline, an MLP residual block requires 33ms, while an attention block requires 16ms. Using the Top-K method of \citet{chern2022tpuknnknearestneighbor}, the sparse MLP block takes 89ms, which is approximately 2.7$\times$ slower than dense MLP. This overhead causes the Top-K operation to dominate not only the MLP computation but also the overall transformer block (MLP + attention $\approx$ 50ms in the dense setting). In contrast, our method increases the sparse MLP latency to only 38 ms, representing a modest +5 ms overhead relative to the dense baseline.
\end{document}